\documentclass[aps,prx,reprint,superscriptaddress]{revtex4-2}
\usepackage{amsmath,amssymb}
\usepackage{bbold}
\usepackage{graphicx}
\usepackage{dcolumn}
\usepackage{bm}
\usepackage[colorlinks]{hyperref}
\usepackage{braket}
\usepackage{rotating}
\usepackage{booktabs}
\usepackage{mathtools}
\usepackage{comment}
\usepackage{xcolor}
\usepackage[title]{appendix}
\usepackage{soul}
\usepackage{color}
\usepackage{physics}
\usepackage{enumerate}
\usepackage{subcaption}
\usepackage{tikz-cd} 
\usepackage{tikz}
\usetikzlibrary{positioning, arrows.meta, shapes.geometric, fit, backgrounds}
\usepackage{cleveref}
\usepackage{algpseudocode}
\usepackage{newfloat}
\DeclareFloatingEnvironment[name=Algorithm,placement=tbp,fileext=loa]{algorithm}
\usepackage{etoolbox}

\DeclareCaptionFormat{ruledalg}{%
  {\hrule height0.9pt}\vskip2pt#1#2#3\par\vskip2pt{\hrule height0.5pt}\vskip3pt}
\AtEndEnvironment{algorithm}{\vskip3pt\hrule height0.9pt}
\AtEndEnvironment{algorithm*}{\vskip3pt\hrule height0.9pt}
\crefmultiformat{equation}{(#2#1#3)}%
{ and~(#2#1#3)}{, (#2#1#3)}{ and~(#2#1#3)}

\colorlet{alena}{blue!80!gray}
\colorlet{roman}{green!40!gray}

\newcommand{\ES}[3]{\mbox{$(#1#2#3)$-ES}}

\begin{document}

\title{LLM-Driven Evolutionary Generation of \\ Multi-Objective Bayesian Optimization Algorithms}

\author{G.~Laskaris}
\affiliation{Terra Quantum AG, Kornhausstrasse 25, 9000 St. Gallen, Switzerland}
\affiliation{LIACS, Leiden University, Leiden, Netherlands}

\author{R.~Brasher}%
\affiliation{Terra Quantum AG, Kornhausstrasse 25, 9000 St. Gallen, Switzerland}

\author{N.~van Stein}%
\affiliation{LIACS, Leiden University, Leiden, Netherlands}

\author{E.~Raponi}%
\affiliation{LIACS, Leiden University, Leiden, Netherlands}

\author{T.~Bäck}%
\affiliation{Terra Quantum AG, Kornhausstrasse 25, 9000 St. Gallen, Switzerland}
\affiliation{LIACS, Leiden University, Leiden, Netherlands}

\author{F.~Neukart}
\affiliation{Terra Quantum AG, Kornhausstrasse 25, 9000 St. Gallen, Switzerland}
\affiliation{LIACS, Leiden University, Leiden, Netherlands}

\begin{abstract}

Designing effective multi-objective Bayesian optimization (MOBO) algorithms requires balancing many interdependent design choices whose optimal configuration is problem-dependent and typically demands deep expertise. We extend the LLaMEA framework to MOBO, using large language models as mutation and crossover operators within evolutionary strategies to generate complete algorithm implementations, with SMAC hyperparameter optimization integrated into the evolutionary loop. Across nine evolutionary runs we generated approximately 900 algorithms and benchmarked them on twelve synthetic problems (ZDT, DTLZ, WFG) and three real-world engineering problems (RE), using a BoFire qParEGO implementation as a state-of-the-art Bayesian-optimization baseline. On the synthetic suite the strongest generated algorithm attains the highest mean normalized hypervolume (0.971, vs. 0.869 for qParEGO) while requiring roughly 60× less wall-clock time; a Friedman test with post-hoc analysis places the two in a single top-performing group, and per-problem tests find the generated algorithm significantly better than qParEGO on 7 of the 12 problems and never worse, matching state-of-the-art accuracy at an order-of-magnitude lower cost. On the three unseen real-world engineering problems a generated algorithm attains the best mean normalized hypervolume (0.985, vs. 0.971 for qParEGO)—significantly better than qParEGO on two of the three problems—at roughly 3.4× lower wall-clock cost, confirming that the gains transfer beyond the synthetic regime. LLM-driven evolutionary search can thus discover algorithm designs that achieve Pareto-efficient trade-offs difficult to reach through manual design.
\end{abstract}

\maketitle

\section{Introduction}\label{sec:introduction}

Multi-objective optimization (MOO) problems arise naturally across science and engineering with extensive applications to drug design~\citep{schneider2020rethinking,stokes2020deep} and structural engineering~\citep{deb2011multi,marler2004survey} among others. When the objective functions are expensive to evaluate, Bayesian optimization (BO)~\citep{shahriari2015taking,jones1998efficient} offers a framework for Pareto-optimal solutions with limited budget. Designing an effective multi-objective Bayesian optimization (MOBO) algorithm remains a challenge since there are numerous interdependent design choices. The best configuration is often problem-dependent, requiring deep expertise and extensive empirical tuning.

Recent work on LLM-driven algorithm generation~\citep{romera2024funsearch, liu2024evolution, ye2024reevo, vanstein2024llamea, novikov2025alphaevolve} has shown that large language models, coupled with evolutionary search, can automatically design optimization algorithms competitive with human-designed alternatives—going beyond parameter tuning to synthesize new algorithms.

LLaMEA-BO~\citep{li2025llamea} is an automated method that uses large language models as variation operators within an evolutionary loop to design complete Bayesian optimization algorithms. In this work, we extend it to the multi-objective setting, with the discovered algorithms as the object of study. We frame the work as an extension of the LLaMEA framework rather than as a new automated-design paradigm: our contribution is the MOBO generation loop and the concrete, competitive algorithms it produces, which we benchmark against established hand-designed MOBO methods. Our contributions are as follows:

\begin{itemize}
\item We extend the LLaMEA framework~\citep{vanstein2024llamea,li2025llamea} to the multi-objective Bayesian optimization setting, generating and configuring complete MOBO algorithms with normalized hypervolume across a suite of synthetic and real-world problems as the fitness signal. Following the LLaMEA-HPO methodology~\citep{van2024loop}, we integrate SMAC-based hyperparameter optimization directly into the evolutionary loop, implemented within our own multi-objective pipeline.
\item We benchmark three evolutionary-strategy configurations within this framework---$(1{+}1)$-ES, $(4{+}16)$-ES, and $(8,16)$-ES---running Gemini-2.5-flash as the LLM operator, and analyze how the selection regime affects the diversity and quality of the generated algorithms.
\item We show that the strongest generated algorithm on the synthetic suite, \textit{MOEAD-EI Hybrid}, attains the highest mean normalized hypervolume across the twelve synthetic problems---significantly more accurate than the state-of-the-art BoFire \textit{qParEGO} baseline on 7 of the 12 problems (per-problem Welch $t$-test) and never significantly worse---while requiring roughly $60\times$ less wall-clock time. On three unseen real-world engineering problems, the best systematically generated algorithm, \textit{Improved-Scalarized-EI}, is the single most accurate method, significantly outperforming \textit{qParEGO} on two of the three problems at $\approx$$3.4\times$ lower cost.
\item We characterize the time--accuracy trade-off across all nine benchmarked algorithms (four LLaMEA-generated, the state-of-the-art \textit{qParEGO} baseline, the surrogate-assisted \textit{IOC-SAMO-COBRA}, two classical evolutionary baselines, and Random Search), showing that the generated algorithms occupy the efficient frontier: they match or exceed state-of-the-art accuracy while running one to three orders of magnitude faster, and every one is significantly faster than \textit{qParEGO} on all fifteen problems (Wilcoxon signed-rank, $p = 6.1\times10^{-5}$).
\end{itemize}

The remainder of this paper is organized as follows. Section~\ref{sec:related} reviews related work in multi-objective Bayesian optimization, LLM-based algorithm design, and automated algorithm configuration. Section~\ref{sec:methodology} describes the LLaMEA-BO framework as extended to the multi-objective setting. Section~\ref{sec:setup} presents the experimental setup---the nine evolutionary runs, the twelve synthetic benchmark problems, and the three real-world engineering problems used for generalization assessment. Section~\ref{sec:results} reports benchmarking results for each of the three phases and the time--accuracy trade-off analysis. Section~\ref{sec:discussion} explains why the selected algorithms perform well and outlines limitations and future directions. Section~\ref{sec:conclusion} concludes.

\section{Related work}\label{sec:related}

\subsection{Multi-Objective Bayesian Optimization}

Bayesian optimization extends to multi-objective settings by replacing the scalar acquisition function with one that accounts for multiple objectives. ParEGO~\citep{knowles2006parego} achieves this through random scalarization: at each iteration it draws a weight vector, scalarizes the objectives via the augmented Tchebycheff decomposition, and maximizes the Expected Improvement (EI) on the resulting scalar surrogate. A limitation of ParEGO is the evaluation of a single scalarization per iteration affecting the simultaneous exploration of different Pareto front regions. qParEGO~\citep{daulton2020ehvi} addresses this with a Monte Carlo, parallel reformulation that draws a fresh random scalarization for each point in a batch, enabling $q$ candidates to be selected jointly per iteration; we adopt its implementation from the BoFire framework~\citep{durholt2024bofire} (specifically its \texttt{QparegoStrategy}, which is built on BoTorch~\citep{balandat2020botorch}) as a state-of-the-art baseline in our experiments.

An alternative family of methods directly targets the hypervolume indicator. Expected Hypervolume Improvement (EHVI)~\citep{emmerich2006ehvi} 
computes the expected gain in dominated hypervolume from a candidate point, providing an acquisition function that balances convergence and diversity. The limitation whereby EHVI computations scale poorly with the number of objectives has been mitigated by Monte Carlo approximations~\citep{daulton2020ehvi} and differentiable formulations within the BoTorch framework~\citep{balandat2020botorch}.

Among the information-theoretic approaches are MESMO~\citep{belakaria2019mesmo} which extends max-value entropy search to the multi-objective setting and PESMO~\citep{hernandez2016pesmo} which uses predictive entropy search to identify points that maximally reduce uncertainty. Decomposition-based methods such as MOEA/D-EGO~\citep{zhang2010moead} combine the weight-vector decomposition strategy from evolutionary multi-objective optimization with Gaussian process surrogates.

Despite the handful of approaches, each algorithm embodies specific design decisions---surrogate type, acquisition function, candidate generation, normalization---that are difficult to recombine or modify without expert knowledge. Our work avoids this manual design process utilizing an LLM to generate MOBO algorithms from scratch.

\subsection{LLM-Based Algorithm Design}

The use of LLMs for automated algorithm design was opened by FunSearch~\citep{romera2024funsearch}, an evolutionary procedure showing that an LLM guided by execution feedback can discover novel mathematical constructs and new solutions to the cap-set problem~\citep{grochow2019new}. Shortly afterwards, LLaMEA~\citep{vanstein2024llamea} introduced a framework in which an LLM acts as the variation operator within an evolution strategy to generate single-objective metaheuristics, receiving the current best algorithm together with its performance feedback and returning an improved variant. The paradigm was later scaled to a broad class of scientific and engineering problems by AlphaEvolve~\citep{novikov2025alphaevolve}, which couples an ensemble of frontier LLMs with an evolutionary outer loop and an automated evaluation harness, reporting discoveries such as improved matrix-multiplication algorithms; open re-implementations (OpenEvolve~\citep{openevolve}) and related systems (ShinkaEvolve~\citep{lange2025shinkaevolve}) have followed. A parallel line targets automatic \emph{heuristic} design: Evolution of Heuristics (EoH)~\citep{liu2024evolution} co-evolves natural-language design ideas and their code, and ReEvo~\citep{ye2024reevo} adds reflection on past failures to steer generation, while other methods replace the evolutionary population altogether---LHNS~\citep{xie2025llm} with LLM-driven neighborhood search and MCTS-AHD~\citep{zheng2025monte} with Monte-Carlo tree search. EvoPrompting~\citep{chen2023evoprompting} applies a comparable evolutionary LLM loop to neural-architecture search. These methods target different problem classes and few direct comparisons exist across the families, so we treat them as context rather than as baselines.

Most relevant to our work, LLaMEA-BO~\citep{li2025llamea} extended LLaMEA to single-objective Bayesian optimization, generating BO algorithms with their own surrogate models and acquisition functions, and LLaMEA-HPO~\citep{van2024loop} integrated SMAC-based hyperparameter optimization into the evolutionary loop so that each generated algorithm's tunable parameters are configured automatically.

We build on this lineage---rather than the heuristic-design methods above---because it is purpose-built for our setting: to our knowledge, LLaMEA-BO is the only LLM-driven framework that synthesizes \emph{complete} BO algorithms. Because its modular design decouples the problem suite and evaluator from the generation loop, extending it to the multi-objective setting is primarily a matter of changing the evaluation signal---to normalized hypervolume over a Pareto-front approximation---rather than redesigning the generation mechanism.

\subsection{Automated Algorithm Configuration}

Automated algorithm configuration seeks to find optimal parameter settings for a given algorithm across a set of problem instances. SMAC~\citep{lindauer2022smac} uses a random forest surrogate to model the relationship between configurations and performance, employing Bayesian optimization to efficiently search the configuration space. irace~\citep{lopez2016irace} takes an iterated racing approach, progressively eliminating poor configurations through statistical testing.

Our work builds directly on LLaMEA-BO and LLaMEA-HPO, extending them to the multi-objective setting. This extension is non-trivial: multi-objective algorithms must handle Pareto dominance, hypervolume computation, objective normalization, and decomposition strategies that have no counterpart in the single-objective case.

\section{Methodology}\label{sec:methodology}

\subsection{Framework Overview}

Our multi-objective implementation LLaMEA-MOBO, operates as an evolutionary loop in which an LLM serves as both the mutation and crossover operator. Figure~\ref{fig:framework} illustrates the pipeline. At each generation, the framework: 

\begin{figure*}[t]
  \centering
  \includegraphics[width=\textwidth]{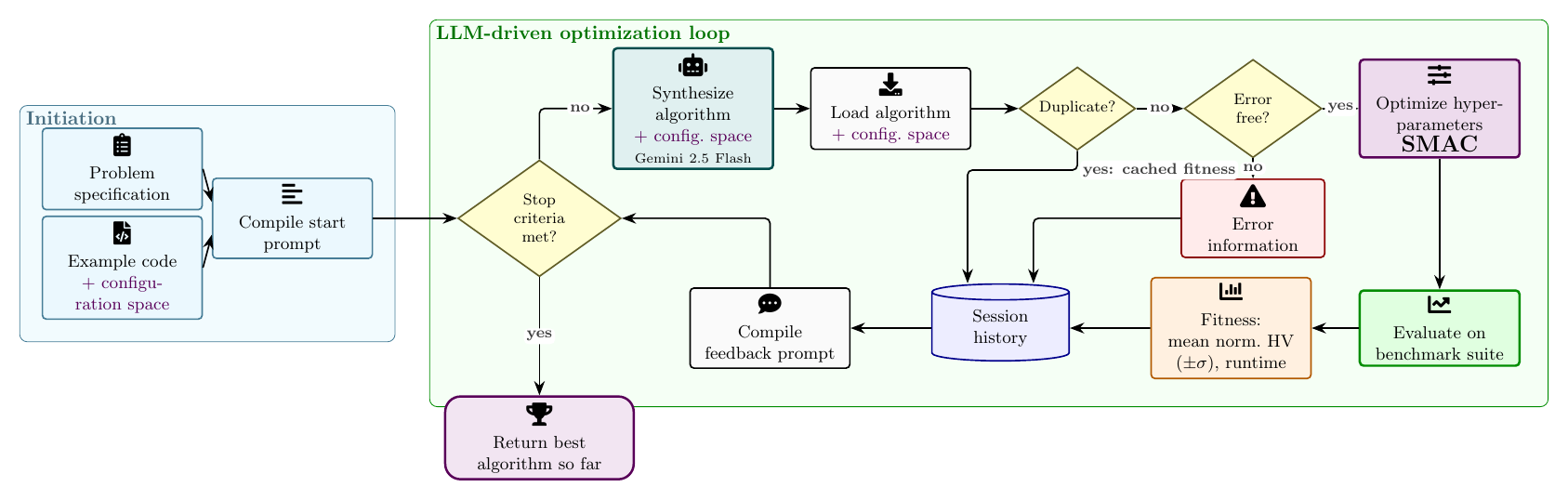}
  \caption{Overview of the LLaMEA-MOBO framework. An \emph{initiation} stage compiles
  the start prompt from a problem specification and an example algorithm
  (with its configuration space). The \emph{LLM-driven optimization loop} then
  synthesizes candidate algorithms with Gemini~2.5~Flash, filters duplicates and
  non-compiling code, tunes hyper-parameters with SMAC, evaluates each candidate
  on the benchmark suite, and feeds the resulting fitness (and any error
  information) back through the session history until the stopping criterion is met.}
  \label{fig:framework}
\end{figure*}

\begin{enumerate}
\item \textbf{Prompts the LLM} with the parent algorithm's source code, its fitness feedback (normalized hypervolume with one standard deviation and execution time), and any runtime errors from the previous evaluation. For mutation, a single parent is provided with the instruction to \emph{refine the strategy of the selected solution to improve it}; for crossover, two parents are provided with the instruction to \emph{combine the selected solutions into a new solution and then refine it}, additionally analyzing and avoiding any runtime errors reported for the parents.
\item \textbf{Extracts the response}, which must follow a structured format: a natural-language description, a justification of design choices, the complete Python implementation, and a hyperparameter search space definition.
\item \textbf{Detects duplicates} by stripping comments and comparing against all previously generated algorithms using \texttt{difflib.SequenceMatcher} with a 0.99 similarity threshold. Duplicates reuse cached fitness values.
\item \textbf{Optimizes hyperparameters} via SMAC (Section~\ref{sec:hpo}), tuning the algorithm's configurable parameters across multiple problem instances.
\item \textbf{Evaluates fitness} as the normalized hypervolume averaged across benchmark problems, and feeds the result back to the LLM for the next generation.
\end{enumerate}

Each LLM-generated algorithm must conform to a fixed object-oriented interface. The constructor receives the evaluation budget, decision-space dimensionality, and variable bounds together with any tunable hyperparameters, while a separate callable entry point executes the optimization loop given a black-box objective function and returns its final approximation of the Pareto front. Crucially, the number of objectives is not supplied as an input: algorithms must infer it at runtime from the first function evaluation, which ensures generality across bi-objective and many-objective problems without manual reconfiguration.

\subsection{Evolutionary Strategies}

We study several evolutionary strategy configurations. The \ES{\text{$\mu$}}{+}{\text{$\lambda$}}~\citep{beyer2001theory} is an elitist selection strategy where off-springs are produced through mutation (single parent) or crossover (two or more parents). In particular, we examine the \ES{1}{+}{1} and \ES{4}{+}{16} cases. The first, maintains a single parent and generates one offspring per generation, which leads to a sequential refinement of the generated algorithms. The second strategy, maintains a population of $\mu = 4$ parents and in each generation produces $\lambda = 16$ off-springs. Under elitist, the best $4$ individuals from the combined parent-offspring pool survive to the next generation. This configuration favors the exploration of diverse generated algorithms, trading-off the exploitative sequential refinement of \ES{1}{+}{1}. We examine also a non-elitist evolutionary strategy \ES{\text{$\mu$}}{,}{\text{$\lambda$}}~\citep{beyer2001theory} with $\mu = 8$ and $\lambda = 16$. This strategy allows for more exploration and smaller probability of getting stuck in a local optima but also means slower convergence. Finally, across all three configurations, parents are ranked by fitness (mean normalized hypervolume) and offspring are generated from this ranked set with reuse permitted: each offspring applies crossover with probability $0.6$ and mutation otherwise, drawing two parents for crossover and a single parent for mutation. The $(1{+}1)$-ES, which maintains a single parent, therefore relies on mutation alone.

\subsection{Hyperparameter Optimization}
\label{sec:hpo}

Each generated algorithm defines a set of tunable hyperparameters (e.g., batch size, kernel parameters, number of weight vectors) together with a configuration space specifying their admissible ranges and types. The LLM is instructed to emit this configuration space alongside the algorithm code, as a Python dictionary whose keys are hyperparameter names and whose values encode the parameter type and search range: a two-tuple $(\ell, u)$ denotes a continuous or integer range between bounds $\ell$ and $u$ with uniform sampling; a three-tuple $(\ell, u, \texttt{"log"})$ denotes a log-uniform range, used for quantities such as learning rates or regularization strengths that span several orders of magnitude; and a list of values denotes a categorical choice. To keep the search tractable and the configurations comparable across candidates, the LLM is constrained to declare between three and seven hyperparameters per algorithm, to use log-scaled ranges where appropriate, and to exclude problem-specific parameters such as budget, dimensionality, and bounds, which are fixed by the benchmark and not optimized. The resulting dictionary is parsed into a configuration-space object compatible with the SMAC hyperparameter optimizer used in the next step. 

Following the LLaMEA-HPO approach~\citep{van2024loop}, we use
SMAC~\citep{lindauer2022smac} with multi-fidelity optimization to search the
hyperparameter space. As in that work, each configuration is evaluated across
an evenly-spaced subset of the benchmark suite to promote generalization
rather than overfitting to a single problem. The objective minimized by SMAC is $1 - \bar{h}$, where $\bar{h}$ is the mean normalized hypervolume across instances, defined as:
\begin{equation}
\bar{h} = \frac{1}{|P|} \sum_{p \in P} \frac{\mathrm{HV}(F_p, r_p)}{\prod_{i} r_{p,i}}
\end{equation}
where $F_p$ is the Pareto front approximation on problem $p$, $r_p$ is the reference point, and $\mathrm{HV}(\cdot)$ is the hypervolume indicator. Dividing by $\prod_i r_{p,i}$ normalizes across problems with different objective scales, ensuring that no single problem dominates the fitness signal.

Before running the full SMAC optimization, we test the generated algorithm with a random configuration and a reduced budget, to detect any errors and avoid running SMAC on a non-functional code. The incumbent configuration found by SMAC is then used for the final fitness evaluation on the complete benchmark suite.

\section{Experimental setup}\label{sec:setup}

\subsection{Algorithms}

We execute a total of nine runs (three \ES{1}{+}{1}, three \ES{4}{+}{16}, and three \ES{8}{,}{16}). The budget is defined as the number of algorithms evaluated across all complete generations rounded up to reach at least 100. This yields 100 evaluated algorithms for \ES{1}{+}{1} and \ES{4}{+}{16} (4 initial parents + 6 generations × 16 offspring) and 104 for \ES{8}{,}{16} (8 initial parents + 6 generations × 16 offspring). From each run we select the best-performing generated algorithm for benchmarking. Additionally, we compare against a pool of baselines spanning a state-of-the-art Bayesian-optimization method (\textit{qParEGO}~\citep{daulton2020ehvi}), a surrogate-assisted optimizer (\textit{IOC-SAMO-COBRA}~\citep{ioc-samo-cobra}), two evolutionary algorithms (\textit{NSGA-II}~\citep{nsga-ii}, \textit{NSGA-III}~\citep{nsga-iii}), and multi-objective Random Search. For \textit{qParEGO} we use the implementation provided by the BoFire framework~\citep{durholt2024bofire} (built on BoTorch~\citep{balandat2020botorch}), ensuring a faithful and maintained state-of-the-art baseline.

\subsection{Benchmark Problems}

We evaluate all algorithms (both LLaMEA-generated and established baselines) across three phases. Phase~1 ranks the nine LLaMEA-generated algorithms against one another; Phase~2 compares the top three of them---together with the development-found MOEAD-EI Hybrid---against five established baselines; and Phase~3 evaluates those nine algorithms on unseen real-world problems. The first two phases use the synthetic suite; the third uses real-world engineering problems. Synthetic benchmark suites are the de facto standard in the MOO community because they offer known Pareto fronts, controllable difficulty axes (multimodality, non-convexity, disconnectedness, deceptiveness, bias), and reproducible conditions across studies~\citep{zitzler2000comparison,deb2005scalable,huband2006review}. We adopt three such suites and complement them with a curated set of real-world engineering problems to assess generalization.

All problems are treated as black-box, with a fixed budget of 400 expensive function evaluations per run and 5 independent repetitions per algorithm--problem pair to account for stochasticity. Performance is measured by the hypervolume (HV) indicator~\citep{zitzler1999multiobjective, zitzler2003performance}, computed at each evaluation step relative to a fixed reference point.

The synthetic suite combines twelve problems drawn from three established families, all sourced from the \texttt{pymoo} library~\citep{pymoo}:
\begin{itemize}
  \item \textbf{ZDT}~\citep{zitzler2000comparison}: bi-objective problems with
        controllable convexity, multimodality, and connectedness of the
        Pareto front. We include \emph{ZDT1} (convex front), \emph{ZDT2}
        (concave), \emph{ZDT3} (disconnected), \emph{ZDT4} (highly
        multimodal), and \emph{ZDT6} (non-uniform Pareto density).
  \item \textbf{DTLZ}~\citep{deb2005scalable}: a scalable suite designed
        for arbitrary numbers of objectives and decision variables, with
        parameterized features such as multimodality (DTLZ1, DTLZ4) and
        disconnected Pareto fronts (DTLZ7). We use the tri-objective
        configurations \emph{DTLZ1, DTLZ2, DTLZ4}, and \emph{DTLZ7}.
  \item \textbf{WFG}~\citep{huband2006review}: problems featuring bias,
        parameter dependency, and non-separability, providing a stiffer
        test than ZDT/DTLZ. We include \emph{WFG4} and \emph{WFG9}
        (bi-objective) and \emph{WFG7} (tri-objective).
\end{itemize}

The selected twelve problems span decision-space dimensionalities from 5 to 30 variables and both bi- and tri-objective settings. Complete dimensions, objective counts, and reference points used for the HV calculation are listed in Table~\ref{tab:problems}.

For the real-world generalization phase, we use the three unconstrained problems of the RE benchmark suite~\citep{reproblems}: \emph{RE21} (a four-bar truss design balancing weight and deflection), \emph{RE34} (a rocket-injector design with three competing performance objectives), and \emph{RE37} (a gear-train design under three accuracy and weight criteria), summarized in Table~\ref{tab:problems}. The RE suite is curated from peer-reviewed engineering case studies and exhibits the irregular, non-smooth objective geometry that synthetic benchmarks typically lack, making it a natural test for whether algorithms developed
against synthetic problems transfer to genuine deployment scenarios.

\subsection{Benchmark Phases}

In the first phase, we benchmark the best-performing generated algorithm from each LLaMEA run against the synthetic problems listed in Table~\ref{tab:problems}. This suite is designed to be diverse, covering both low- and high-dimensional decision spaces, as well as bi- and tri-objective settings. 
Importantly, only a subset of these problems---ZDT1--4, ZDT6, and DTLZ1-2,4 problems---were used during the LLaMEA generation phase; the remaining problems are held out and unseen by the algorithms during evolution, allowing us to assess generalization.

\begin{table}[t]
    \centering
    \caption{Benchmark problems: twelve synthetic problems (ZDT, DTLZ, WFG; used
    in Phases~1--2) and three real-world engineering problems (RE
    suite~\citep{reproblems}; Phase~3). $d$: decision-space dimension; $m$: number
    of objectives. For the RE problems, reference points are set at $1.1\times$ the
    95th percentile of $10{,}000$ uniform random samples per objective.}
    \label{tab:problems}
    \begin{tabular}{lccc}
        \toprule
        \textbf{Problem} & \textbf{d} & \textbf{m} & \textbf{Reference point} \\
        \midrule
        \multicolumn{4}{l}{\textit{Synthetic (Phases 1--2)}} \\
        ZDT1  & 30 & 2 & $(1.1,\ 7.2)$ \\
        ZDT2  & 30 & 2 & $(1.1,\ 8.0)$ \\
        ZDT3  & 30 & 2 & $(1.1,\ 7.2)$ \\
        ZDT4  & 10 & 2 & $(1.1,\ 300.0)$ \\
        ZDT6  & 10 & 2 & $(1.1,\ 11.0)$ \\
        DTLZ1 &  5 & 3 & $(10.0,\ 10.0,\ 10.0)$ \\
        DTLZ2 & 20 & 3 & $(3.0,\ 3.0,\ 3.0)$ \\
        DTLZ4 &  5 & 3 & $(1.1,\ 1.1,\ 1.1)$ \\
        DTLZ7 & 20 & 3 & $(1.1,\ 1.1,\ 28.0)$ \\
        WFG4  &  6 & 2 & $(2.2,\ 4.4)$ \\
        WFG7  & 10 & 3 & $(2.2,\ 4.4,\ 6.6)$ \\
        WFG9  &  6 & 2 & $(2.2,\ 4.4)$ \\
        \midrule
        \multicolumn{4}{l}{\textit{Real-world RE (Phase 3)}} \\
        RE21 & 4 & 2 & $(2851.9,\ 0.037)$ \\
        RE34 & 5 & 3 & $(1862.3,\ 12.17,\ 0.196)$ \\
        RE37 & 4 & 3 & $(0.887,\ 0.913,\ 1.012)$ \\
        \bottomrule
    \end{tabular}
\end{table}

In the second phase, we gather the three best performing generated algorithms from the first phase and benchmark them against all baselines on the same problem set of phase~1. The best algorithms of first phase come after ranking all nine generated algorithms by their mean normalized hypervolume across all phase~1 problems and repeats. Specifically, for each problem we normalize the final-epoch HV of each algorithm by the maximum HV achieved by any algorithm on that problem, ensuring that problems of vastly different scales (e.g. ZDT4 vs. ZDT1) contribute equally to the ranking. We emphasize that this is a \emph{within-phase} relative score: the per-problem maximum is taken over the algorithms present in each phase and recomputed separately for Phase~1 (the nine generated algorithms) and Phases~2--3 (the generated algorithms together with the baselines). It is therefore distinct from the reference-point-based search fitness of Section~\ref{sec:hpo}, and the normalized HV of a given algorithm is not directly comparable across phases---for instance, Improved-Scalarized-EI reads $0.855$ in Phase~1 but $0.811$ in Phase~2, as the strong baselines raise the per-problem maxima. The three best algorithms from this procedure are Improved-Scalarized-EI, RF-LCB-PBI, and RF-ParEGO-Batch (see Appendix~\ref{app:algorithms}), achieving mean normalized HVs of 0.855, 0.795, and 0.756 respectively.

The third and final phase benchmarks the same algorithms from phase~2 but this time on real-world unconstrained multi-objective problems of RE benchmark suite~\citep{reproblems} as seen in Table~\ref{tab:problems}. Reference points for RE problems are set to $1.1 \times$ the 95th percentile of each objective, estimated from $10,000$ uniformly sampled candidate solutions.

\section{Results}\label{sec:results}

\subsection{LLaMEA-Generated Algorithms}

Across the nine evolutionary runs, LLaMEA generated a total of approximately 900 candidate algorithms. These algorithms span a broad range of surrogate models and acquisition functions, including Gaussian Process-based surrogates with decomposition strategies~\citep{bajer2019gaussian,frazier2018tutorial,lu2023surrogate}, tree-ensemble regressors (random forests and gradient boosting) ~\citep{bhattacharjee2024bayesian,lai2025coffeeboost}, windowed model updates, and hybrid evolutionary-Bayesian acquisition schemes~\citep{wang2023recent,li2023evolutionary}. This diversity of strategies demonstrates that LLaMEA does not favor any particular algorithmic type but rather explores different regions of the algorithmic design space.

\begin{figure*}[t]
  \centering
  \includegraphics[width=.8\textwidth]{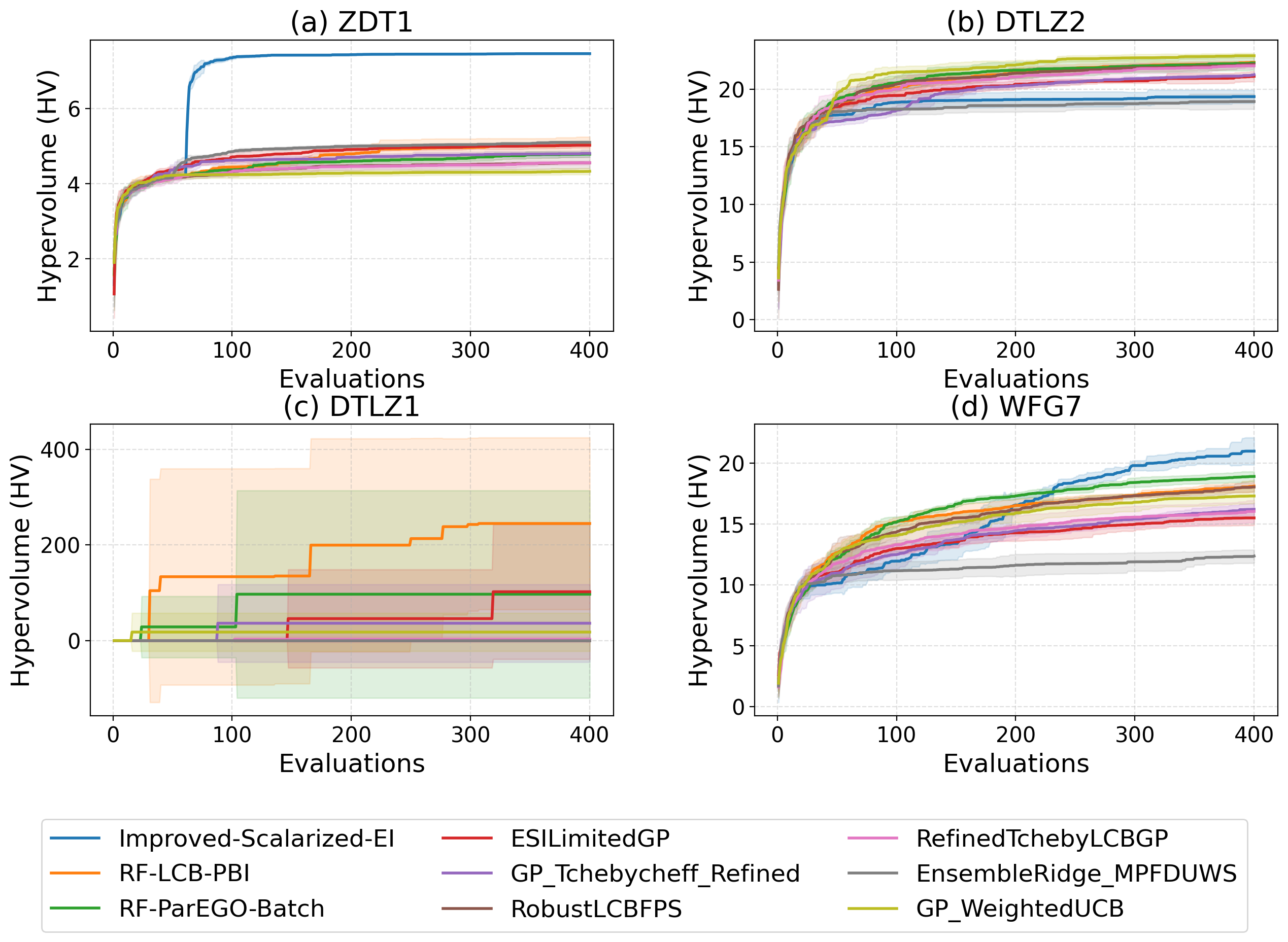}
  \caption{Convergence of the nine LLaMEA-generated algorithms (Phase 1) on
  representative problems from each family. Shaded bands denote $\pm1$ std over
  five repeats.}
  \label{fig:phase1_conv}
\end{figure*}

\begin{figure*}[t]
  \centering
  \includegraphics[width=\textwidth]{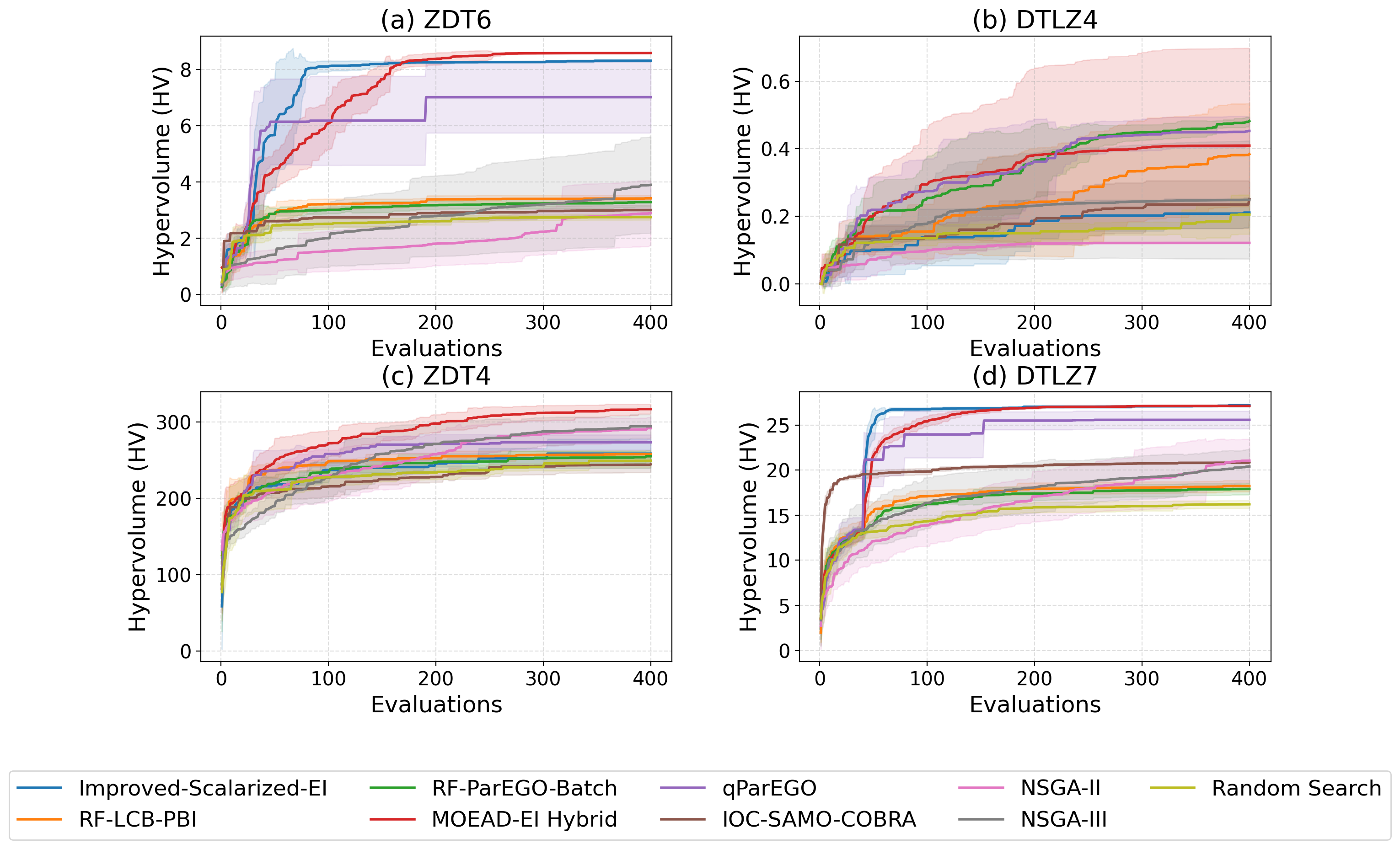}
  \caption{Convergence of all nine algorithms---the four LLaMEA-generated (the
top-3 plus MOEAD-EI Hybrid) and five baselines/SOTA---on representative Phase-2
problems. Shaded bands denote $\pm1$ std over five repeats.}
  \label{fig:phase2_conv}
\end{figure*}

As described in Section~\ref{sec:setup}, the three best algorithms are selected by mean normalized HV across all Phase~1 problems. The ranking is as follows: Improved-Scalarized-EI (0.855), RF-LCB-PBI (0.795), and RF-ParEGO-Batch (0.756); we examine these designs and the mechanisms behind their performance in Section~\ref{sec:top3}. In addition to the systematically selected algorithms, during the development of our experimentation we discovered MOEAD-EI Hybrid, an algorithm that proved the strongest overall on the synthetic benchmark. Given its exceptional performance, we retain it as a generated candidate throughout the benchmark comparison and analyze it alongside the selected algorithms in Section~\ref{sec:top3}. 

\subsection{Benchmarking Results}

\subsubsection{Phase 1: Comparison Among Generated Algorithms}

In Phase~1, the nine best-performing generated algorithms are evaluated on the twelve synthetic benchmark problems listed in Table~\ref{tab:problems}, using a budget of 400 function evaluations and 5 independent repeats per problem. Results are summarized in terms of HV convergence curves and final normalized HV.

Improved-Scalarized-EI achieves the highest mean normalized HV (0.855), winning 8 of the 12 problems: it dominates the entire ZDT family and also leads on DTLZ7, WFG4, and WFG7. On ZDT4 — a highly multimodal problem — it reaches a mean final HV of 258.4, the highest among the generated algorithms (against 218–257 for the others). Its advantage is not uniform, however: the random-forest-based RF-LCB-PBI and RF-ParEGO-Batch lead on the harder DTLZ problems (DTLZ1, DTLZ4), where Improved-Scalarized-EI is markedly less effective, reflecting complementary strengths across the generated set.

All nine generated algorithms complete all twelve problems given an adequate per-run time budget; the substantial differences in their computational cost are analyzed in Section~\ref{sec:time_accuracy}.

\subsubsection{Phase 2: Comparison Against Baselines and State-of-the-Art on Synthetic Problems}

In Phase~2, the best three generated algorithms — together with the development-found MOEAD-EI Hybrid — are benchmarked alongside five established algorithms — Multi-objective Random Search, NSGA-II~\citep{nsga-ii}, NSGA-III~\citep{nsga-iii}, IOC-SAMO-COBRA~\citep{ioc-samo-cobra}, and qParEGO~\citep{daulton2020ehvi} — on the same twelve synthetic problems under identical conditions.

\begin{table}[t]\centering
\caption{Mean normalized hypervolume ($\pm$ std across problems) in Phase 2 (12 synthetic) and Phase 3 (3 real-world). Bold indicates the best mean per column. $\dagger$ denotes LLaMEA-generated algorithms.}
\label{tab:normalized_hv}
\begin{tabular}{lcc}\toprule
Algorithm & Phase 2 & Phase 3 \\ \midrule
Improved-Scalarized-EI$^\dagger$ & $0.811\pm0.301$ & $\mathbf{0.985\pm0.026}$ \\
RF-LCB-PBI$^\dagger$ & $0.700\pm0.183$ & $0.853\pm0.008$ \\
RF-ParEGO-Batch$^\dagger$ & $0.693\pm0.242$ & $0.902\pm0.048$ \\
MOEAD-EI Hybrid$^\dagger$ & $\mathbf{0.971\pm0.047}$ & $0.926\pm0.053$ \\
qParEGO & $0.869\pm0.234$ & $0.971\pm0.038$ \\
IOC-SAMO-COBRA & $0.699\pm0.248$ & $0.795\pm0.060$ \\
NSGA-II & $0.691\pm0.233$ & $0.849\pm0.023$ \\
NSGA-III & $0.765\pm0.162$ & $0.842\pm0.082$ \\
Random Search & $0.600\pm0.231$ & $0.796\pm0.063$ \\
\bottomrule\end{tabular}\end{table}

MOEAD-EI Hybrid achieves the highest mean normalized HV of 0.971, beating qParEGO on 8 of the 12 problems and outperforming every classical baseline; in the per-problem analysis it is significantly more accurate than the state-of-the-art qParEGO (0.869) on 7 of the 12 problems and never significantly worse (Section~\ref{subsec:significance}). Among the systematically selected algorithms, Improved-Scalarized-EI is the strongest at 0.811 — ahead of every classical baseline (best: NSGA-III, 0.765) and behind only qParEGO (0.869) and MOEAD-EI Hybrid (0.971) — whereas the random-forest-based RF-LCB-PBI (0.700) and RF-ParEGO-Batch (0.693) perform on par with the classical baselines on the synthetic suite. NSGA-II performs competitively at 0.691, reflecting the well-known effectiveness of evolutionary methods on synthetic benchmarks with moderate budget. The classical baselines (NSGA-II, NSGA-III, IOC-SAMO-COBRA, Random Search) require negligible computation — under one second per repeat — but rank below the strongest generated algorithm, confirming that surrogate-assisted search provides meaningful gains within a budget of 400 evaluations. Random Search achieves a mean normalized HV of 0.600, establishing the lower bound of the comparison.

\subsubsection{Phase 3: Generalization to Real-World Problems}

Phase~3 evaluates all nine algorithms on three unconstrained real-world engineering optimization problems from the RE benchmark suite~\citep{reproblems} (Table~\ref{tab:problems}), which were unseen during the LLaMEA generation phase and during Phase~1 and Phase~2 benchmarking.

\begin{figure}[t]
  \centering
  \includegraphics[width=\columnwidth]{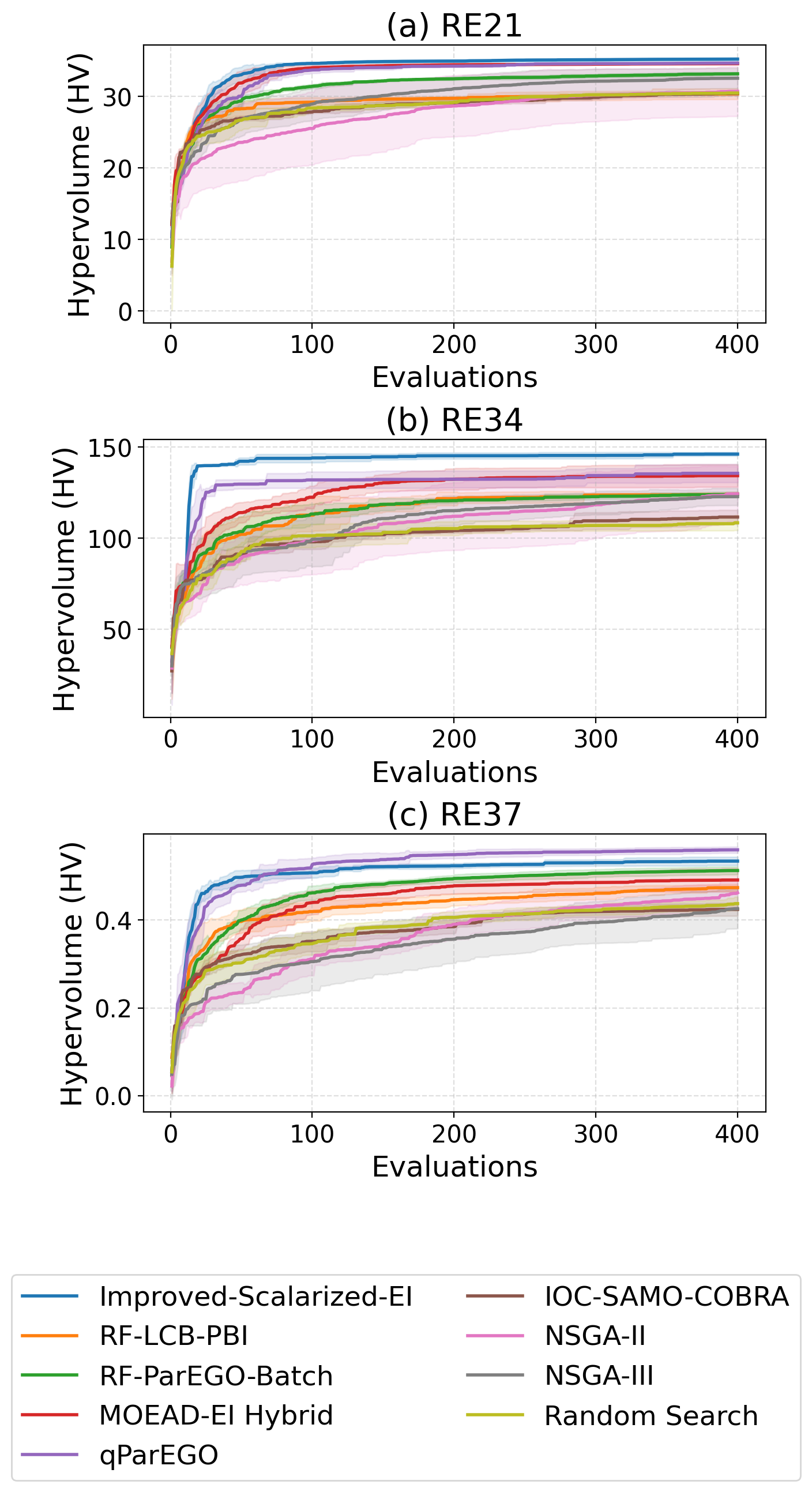}
  \caption{Convergence on the three real-world RE problems (Phase 3) for all
  nine algorithms. Shaded bands denote $\pm1$ std over five repeats.}
  \label{fig:phase3_conv}
\end{figure}

Here the ordering changes. Improved-Scalarized-EI attains the highest mean normalized HV (0.985), narrowly ahead of the state-of-the-art qParEGO (0.971); MOEAD-EI Hybrid follows at 0.926---third overall, ahead of RF-ParEGO-Batch (0.902) and every classical baseline---so two generated algorithms join qParEGO at the top of the real-world ranking. Notably, NSGA-II remains competitive (0.849) despite its complete absence of a surrogate model, consistent with prior findings that population diversity is particularly valuable in small-dimensional, multi-peaked landscapes. As the real-world suite contains only three problems, we report these results descriptively; the rank-based test over it is underpowered (Section~\ref{subsec:significance}).

\subsubsection{Time--Accuracy Trade-off}
\label{sec:time_accuracy}

A central contribution of this work is the identification of LLaMEA-generated algorithms that match or exceed state-of-the-art accuracy at a fraction of the computational cost. Wall-clock time is measured as the total execution time for one complete optimization run of 400 function evaluations on a single problem instance, averaged across all problems and independent repetitions within the respective benchmark phase. All benchmark runs were executed in parallel across $10$ CPU cores using a process-pool executor; in a controlled comparison against isolated single-process execution, parallel timings were inflated by approximately 30\% due to memory-bandwidth contention and Python global-interpreter-lock overhead. We report parallel timings throughout, as they reflect the practical deployment scenario; the relative ranking between algorithms is unaffected by this inflation, since all algorithms are subject to the same contention. Each algorithm is timed in its own pass over the problem suite, so every method has the full ten-core pool available during its measurement and the timings are directly comparable across algorithms. 

\begin{figure*}[t]
  \centering
  \includegraphics[width=\textwidth]{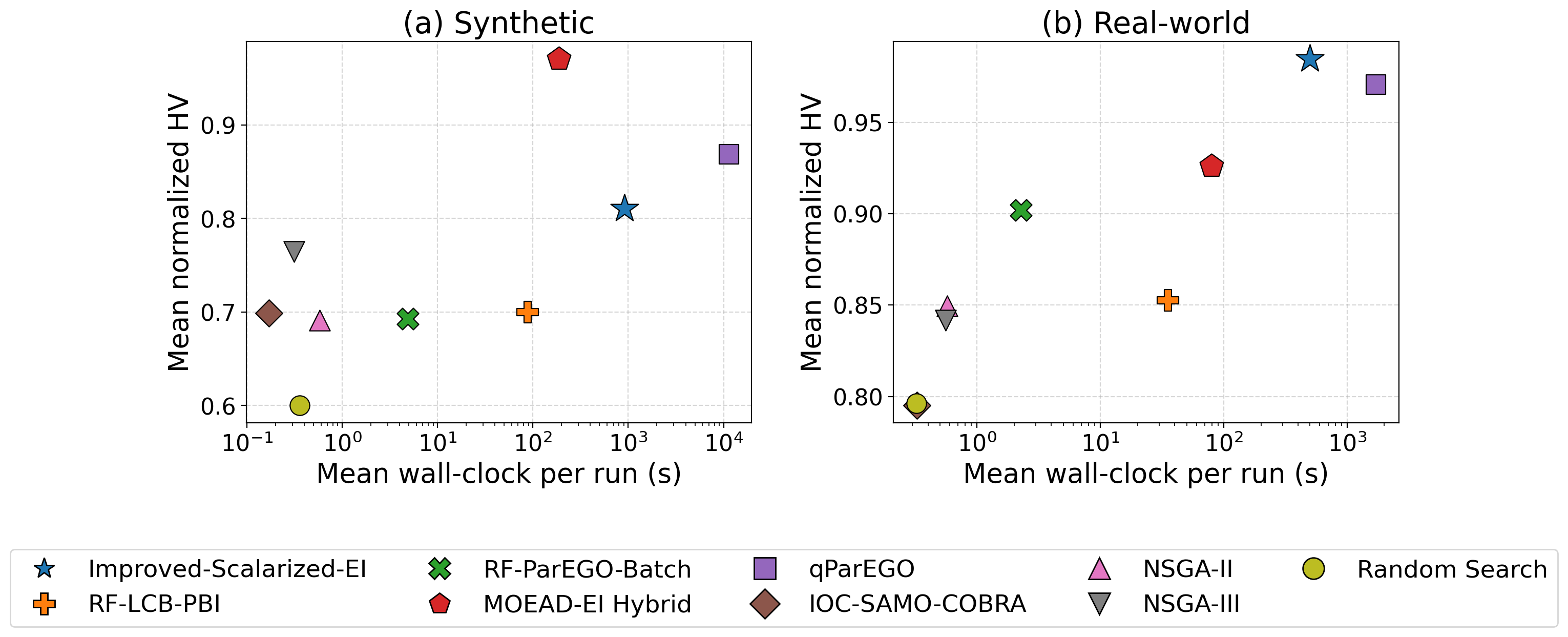}
  \caption{Time–accuracy trade-off on (a) the twelve synthetic and (b) the three real-world (RE) problems. Horizontal axis: mean wall-clock time per run (log scale); vertical axis: mean normalized hypervolume. The LLaMEA-generated algorithms occupy the efficient region.}
  \label{fig:time_accuracy}
\end{figure*}

Figure~\ref{fig:time_accuracy} plots mean normalized HV against mean runtime per repeat for both Phase~2 and Phase~3, making the efficiency landscape explicit.

Three tiers emerge clearly. First, the classical baselines (\textit{NSGA-II}, \textit{NSGA-III}, \textit{IOC-SAMO-COBRA}, \textit{Random Search}) operate at under one second per repeat but achieve moderate HV, forming a fast-but-limited group. Second, \textit{qParEGO} is the most expensive method by a wide margin - a mean of 11{,}304 seconds per run on synthetic problems and 1{,}704 seconds on real-world problems. The LLaMEA-generated algorithms occupy the efficient region in between, spanning the accuracy–cost frontier. MOEAD-EI Hybrid matches or exceeds qParEGO's accuracy at a small fraction of the cost: on the synthetic suite it attains the best accuracy of any method (0.971) at only 188 seconds per run---a 60.0$\times$ reduction---and on the real-world problems 0.926 at 79 seconds (21.5$\times$). Improved-Scalarized-EI is the most accurate method on the real-world problems (0.985) at 499 seconds, a 3.4× reduction relative to qParEGO (and 12.4× on synthetic). At the cheap extreme, RF-ParEGO-Batch reaches competitive accuracy (0.902 on real-world, ahead of every classical baseline) at just 2–5 seconds per run — two-to-three orders of magnitude faster than qParEGO (754× on real-world, 2,310× on synthetic).

\subsection{Statistical Significance}
\label{subsec:significance}

\begin{figure*}[t]
  \centering
  \includegraphics[width=0.6\textwidth]{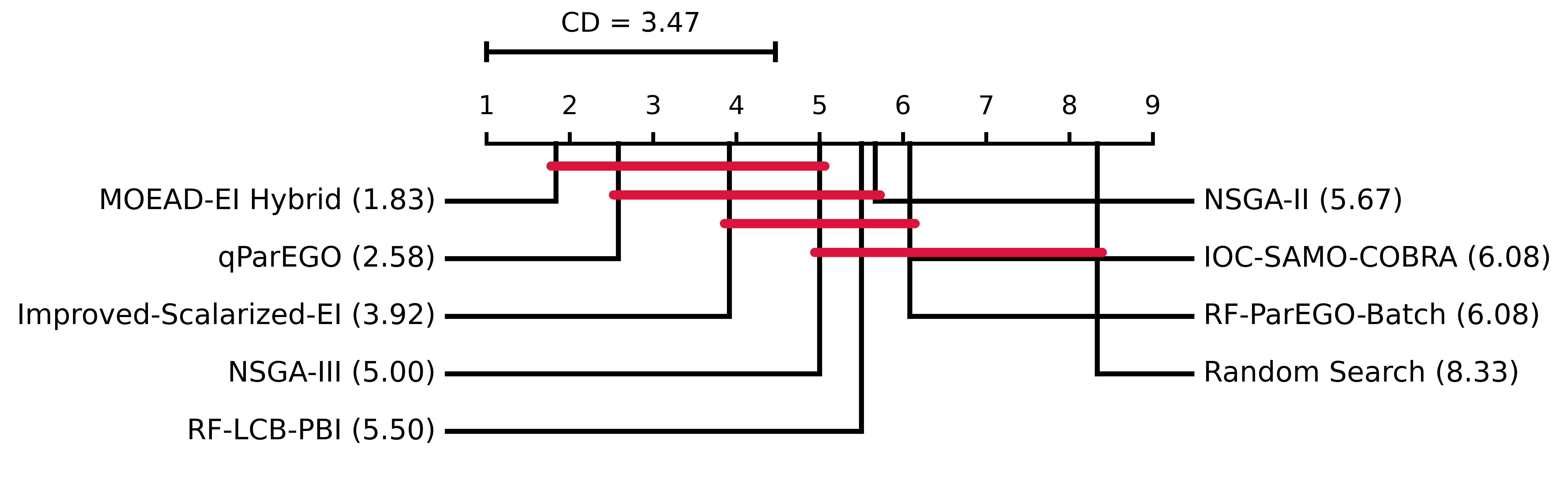}
  \caption{Critical-difference diagram (Nemenyi, $\alpha = 0.05$) of mean ranks across the twelve synthetic problems. Lower rank (left) is better; methods joined by a bar are not significantly different (CD = 3.47). MOEAD-EI Hybrid ranks first (1.83), ahead of qParEGO (2.58) and Improved-Scalarized-EI (3.92).}
\label{fig:cd}
\end{figure*}

Following the per-problem evaluation protocol of the BLADE benchmarking suite~\citep{van2025blade}, we assess significance problem by problem: for each problem a two-sample Welch $t$-test~\citep{welch1947generalization} over the five independent runs ($\alpha=0.05$) compares a generated algorithm against qParEGO, with no multiple-comparison correction (as in BLADE). We focus on the two competitive generated algorithms; each comparison is additionally summarized across the synthetic suite with a Wilcoxon signed-rank test~\citep{wilcoxon1945individual}.

On the synthetic suite, \emph{MOEAD-EI Hybrid} is significantly more accurate than qParEGO on 7 of the 12 problems, never significantly worse, and statistically indistinguishable on the remaining 5; across the suite the improvement is significant (Wilcoxon signed-rank, $p=0.043$). On the three real-world problems it is statistically on par with qParEGO---indistinguishable on re21 and re34 and significantly worse only on re37.

\emph{Improved-Scalarized-EI} matches qParEGO on the synthetic suite: it is significantly better on 4 problems and significantly worse on 4, indistinguishable on the other 4, with no significant suite-level difference (Wilcoxon, $p=0.52$). On the real-world problems, however, it is the single most accurate method, significantly outperforming qParEGO on 2 of the 3 (re21 and re34) and worse only on re37.

The two random-forest variants (RF-LCB-PBI and RF-ParEGO-Batch) are significantly \emph{less} accurate than qParEGO on most problems (10 of 12 synthetic and all three real-world), trading accuracy for one-to-three orders of magnitude lower runtime.

The advantage in computational cost is unequivocal. Every generated algorithm is significantly faster than qParEGO on all 15 problems (Wilcoxon signed-rank over the 15 problems~\citep{wilcoxon1945individual}, $p = 6.1\times10^{-5}$ in every case); the per-phase magnitudes of these reductions are reported in the time–accuracy analysis above.

At the level of the whole suite, a Friedman test~\citep{friedman1937use} on the synthetic problems confirms that the nine algorithms differ significantly ($\chi^2 = 49.91$, $p = 4.3\times10^{-8}$) and ranks MOEAD-EI Hybrid first ($1.83$), ahead of qParEGO ($2.58$) and Improved-Scalarized-EI ($3.92$). The corresponding Nemenyi critical-difference diagram (Fig.\ref{fig:cd}, $\mathrm{CD}=3.47$)~\citep{nemenyi1963distribution, demvsar2006statistical} separates MOEAD-EI Hybrid from random search and the weaker baselines but not from qParEGO; as this all-pairs test has low power on small suites~\citep{demvsar2006statistical}, the per-problem analysis above gives the more direct head-to-head comparison with the state of the art.

\subsection{Effect of Hyperparameter Optimization}\label{sec:hpo_effect}

To isolate the contribution of the SMAC hyperparameter optimization integrated into the generation loop, we compare each of the four highlighted generated algorithms in its deployed SMAC-tuned configuration against the hyperparameters originally proposed by the LLM (its code defaults). Table~\ref{tab:hpo_ablation} reports the mean normalized hypervolume under both settings.

On the synthetic suite---the distribution SMAC tunes on---hyperparameter optimization yields consistent gains for all four algorithms ($+0.9\%$ to $+5.2\%$; $+1.6\%$ for MOEAD-EI Hybrid), largest for the cheaper random-forest designs. On the held-out real-world problems, tuning the three systematically generated algorithms produces no improvement ($-0.3\%$ to $-0.7\%$, within run-to-run variation), whereas for MOEAD-EI Hybrid it yields a large gain ($+7.8\%$, $0.859 \to 0.926$) on a configuration that transfers to the real-world problems. Hyperparameter optimization is thus algorithm-dependent---a modest in-distribution refinement for most designs, and a larger benefit for MOEAD-EI Hybrid's real-world generalization.

\begin{table}[t]\centering
\caption{Effect of SMAC hyperparameter optimization: mean normalized hypervolume with the LLM-proposed default hyperparameters versus the SMAC-tuned incumbent, for the four highlighted LLaMEA-generated algorithms. Normalization uses the same per-problem-max denominator as Table~\ref{tab:normalized_hv}.}
\label{tab:hpo_ablation}
\begin{tabular}{lcccc}
\toprule
 & \multicolumn{2}{c}{Synthetic (12)} & \multicolumn{2}{c}{Real-world (3)} \\
\cmidrule(lr){2-3}\cmidrule(lr){4-5}
Algorithm & default & SMAC & default & SMAC \\
\midrule
Improved-Scalarized-EI & 0.804 & 0.811 & 0.991 & 0.985 \\
RF-LCB-PBI             & 0.684 & 0.700 & 0.859 & 0.853 \\
RF-ParEGO-Batch        & 0.659 & 0.693 & 0.905 & 0.902 \\
MOEAD-EI Hybrid        & 0.956 & 0.971 & 0.859 & 0.926 \\
\bottomrule
\end{tabular}
\end{table}

\section{Discussion}\label{sec:discussion}

\begin{figure}[h!]
    \centering
    \includegraphics[width=\linewidth]{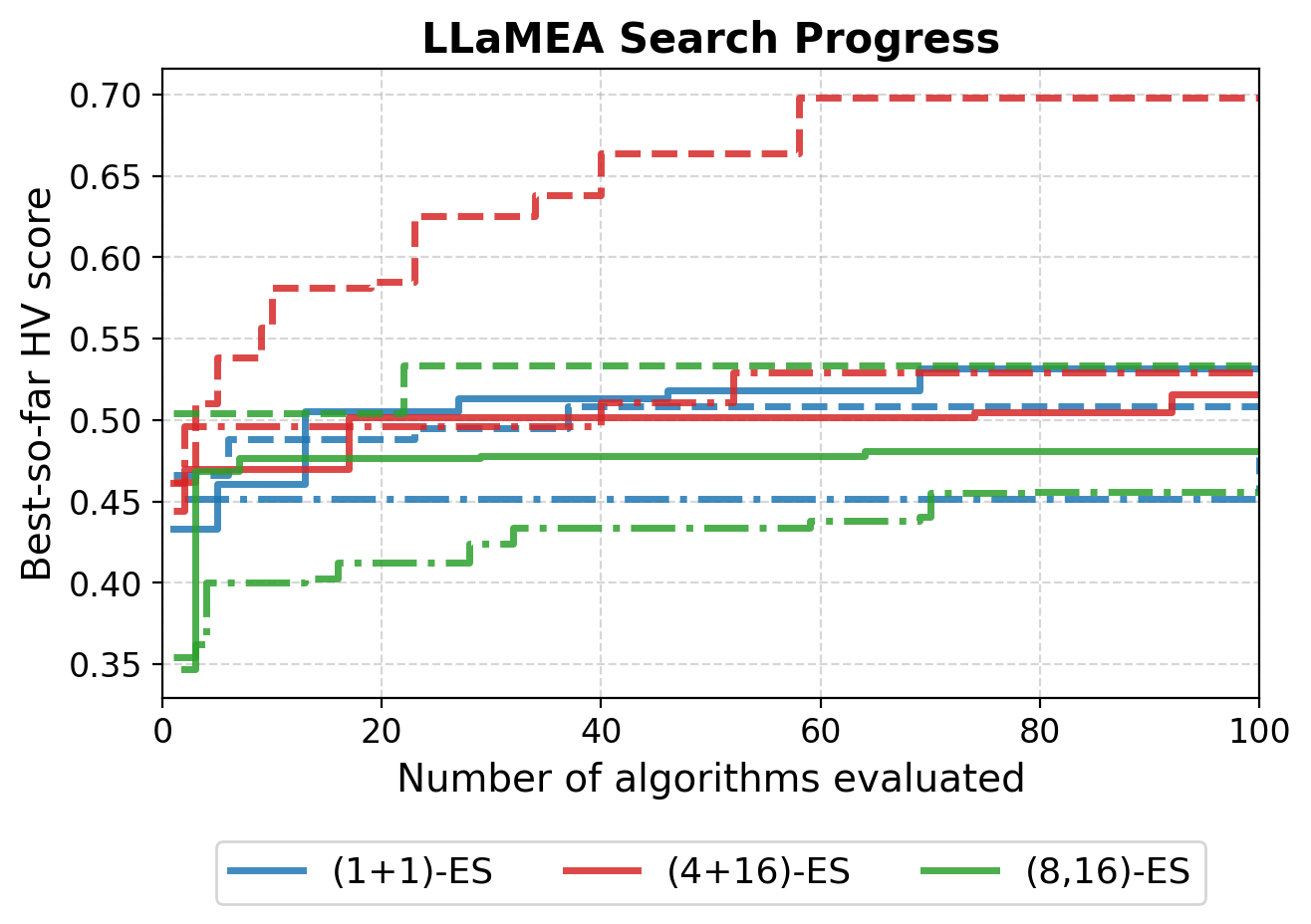}
    \caption{Best-so-far search fitness (mean normalized HV on the training suite) over the first 100 evaluated algorithms, for all nine evolutionary runs (three per evolution strategy; color denotes strategy, line style the run). The \ES{4}{+}{16} runs reach the highest fitness ($\approx 0.70$) and produce the top-ranked Improved-Scalarized-EI, while the \ES{1}{+}{1} and \ES{8}{,}{16} runs plateau lower ($\approx 0.53$).}
\label{fig:search_progress} 
\end{figure}

\subsection{Selected Algorithms Analysis}\label{sec:top3}

To understand not only what the selected algorithms are but how they were discovered, we trace each one back to its origin through the parent links recorded during the search---so the lineages reported here are the actual ancestry of each design, not a post-hoc reconstruction. All four emerge within a single run in at most five generations, each improving markedly over its lineage root (Fig.~\ref{fig:search_progress})---directed refinement rather than one-shot sampling. Despite being generated automatically, all assemble recognizable, well-motivated building blocks, which we relate below to established components of Bayesian and decomposition-based multi-objective optimization.

\begin{itemize}
  \item Reached by mutation over five monotonic generations (search fitness $0.485\rightarrow0.698$; EvolutionaryAcquisition → ESI → ESI-Vec → ESI-ImprovedEA → Improved-Scalarized-EI) that specialize the acquisition and its inner optimizer step by step, \emph{Improved-Scalarized-EI} pairs a Matérn-2.5 Gaussian-process surrogate~\citep{williams2006gaussian} with an Expected-Improvement acquisition~\citep{jones1998efficient} applied to a scale-invariant Tchebycheff scalarization of the objectives~\citep{miettinen1999nonlinear}, and maximizes it with a vectorized evolutionary inner optimizer that uses tournament selection~\citep{eiben2015introduction}. The combination of a calibrated GP posterior and EI's principled exploration--exploitation trade-off is especially effective on the smooth, low-dimensional real-world problems, where it is the single most accurate method; the vectorized EA keeps the acquisition step inexpensive, and a fit-failure fallback makes it robust. 

  \item The best design found during development (search fitness $0.880$), \emph{MOEAD-EI Hybrid} arose at generation 2 by crossover of a fast-but-unstable MOEA/D--GP hybrid and a strong-but-slow ParEGO--GP method, inheriting EI's accuracy while resolving ParEGO's cost. It embeds independent GP surrogates in an MOEA/D decomposition framework~\citep{zhang2007moea}, driving each scalar subproblem with an augmented-Tchebycheff Expected-Improvement acquisition~\citep{steuer1983interactive,miettinen1999nonlinear} and proposing candidates with genetic operators over a fixed set of weight vectors. Decomposition into many scalar subproblems suits the higher-dimensional synthetic suite---where it is the most accurate method overall---and it remains competitive on the low-dimensional real-world problems, ranking third in Phase~3.

  \item RF-LCB-PBI (search fitness $0.489\rightarrow0.530$) swaps the surrogate from $k$-nearest-neighbors to a random-forest ensemble~\citep{breiman2001random,hutter2011sequential} and folds a lower-confidence bound~\citep{srinivas2012information} into the penalty-based boundary-intersection (PBI) scalarization of MOEA/D~\citep{zhang2007moea}. The ensemble variance provides cheap, robust uncertainty estimates that make the method orders of magnitude faster than its GP-based counterparts; the price is a less smooth surrogate, which yields competitive rather than leading accuracy on the continuous synthetic problems.

  \item RF-ParEGO-Batch (search fitness $0.464\rightarrow0.516$) keeps the random-forest surrogate and evolves the acquisition from a multi-objective LCB into a batched ParEGO scalarization~\citep{knowles2006parego}, drawing fresh random scalarization weights each iteration and proposing candidates in batches. This makes it the cheapest method in the study---two to three orders of magnitude faster than qParEGO---deliberately trading surrogate fidelity for throughput while remaining ahead of every classical baseline on the real-world problems.
\end{itemize}

Two broader observations follow. First, the lineages are short and interpretable---every generation corresponds to a named, sensible algorithmic change, with mutation specializing a single module and crossover combining two complementary parents---so the LLM-driven search is legible to a practitioner rather than a black box.

Second, the strongest designs implicitly balance surrogate quality against acquisition cost---for example MOEAD-EI Hybrid's Tchebycheff decomposition with independent GP surrogates, or the random-forest variants' trade of fidelity for speed. This balance is not explicitly optimized by the LLM; it emerges from evaluating candidate algorithms under a fixed evaluation budget, which naturally penalizes slow-converging or computationally expensive designs and rewards those that reach high hypervolume quickly.

\subsection{Limitations}\label{sec:limitations}

Several limitations of the current study should be acknowledged. First, the surrogate-assisted methods differ by orders of magnitude in computational cost: at a $400$-evaluation budget, \textit{qParEGO} alone requires hours per run. This wall-clock expense---not any methodological limit---is what capped the number of repeats and benchmark problems we could include; a broader study would require substantially more compute. Second, the evaluation budget of $400$ function evaluations, while standard for expensive black-box optimization, may not fully reflect the behavior of surrogate-assisted methods at larger scales. Third, the real-world evaluation in Phase~3 is limited to three unconstrained engineering problems; broader conclusions about generalization to constrained, noisy, or higher-dimensional real-world settings require further investigation. Fourth, the quality and diversity of generated algorithms may be sensitive to the choice of LLM backbone and prompt design, aspects that we did not systematically vary.

A further, more subtle hazard concerns \emph{stochasticity}. We did not initially anticipate that, absent an explicit instruction, an LLM may generate \emph{deterministic} heuristics---for instance by fixing the random seed internally. When benchmarking stochastic optimizers this is dangerous: a fixed-seed algorithm reports zero run-to-run variance, which both understates uncertainty and risks overfitting to a single fortunate trajectory. We detected this behavior for \emph{Improved-Scalarized-EI} and addressed it post hoc by re-running it across five independent seeds, restoring a fair variance estimate. A future generation loop should include informed prompts that explicitly require stochastic behavior and proper seed handling, together with an automated validation check that flags zero-variance candidates during the search.

Finally, the development-found MOEAD-EI Hybrid required one mechanical correction before benchmarking: on the lower-dimensional problems its generated code initialized only $n_{\mathrm{init}}$ solutions yet indexed a decomposition population of $N_{\mathrm{pop}}>n_{\mathrm{init}}$ subproblems, raising an out-of-bounds error. We cap the population with a single line, \texttt{self.population\_size = min(self.population\_size, self.n\_init)}, so the algorithm runs consistently across all problem dimensions; the change is purely mechanical and leaves its surrogate and acquisition logic untouched.

\subsection{Future Work}\label{sec:future}

Several directions remain open. Extending LLaMEA to constrained multi-objective problems would substantially broaden the practical applicability of the generated algorithms. Investigating larger evaluation budgets and a wider suite of real-world benchmarks---including noisy and mixed-integer settings---would provide a more complete picture of where LLaMEA-generated algorithms succeed and fail. Furthermore, during the benchmark evaluation phases the discovered algorithms are deployed with the hyperparameter configurations produced during the LLaMEA search; running problem-specific hyperparameter tuning at deployment time (e.g.\ re-invoking SMAC per target problem) could close this gap and further improve their competitiveness on unseen problem instances. Beyond tuning, a further shift is from generating \emph{general-purpose} algorithms---our present goal, where a single discovered algorithm must perform well across a suite---to generating \emph{specialized} solvers for a particular target problem or family. Where evaluations are cheap enough to amortize a dedicated generation run, LLaMEA could be directed at one problem and optimize an algorithm specifically for it; understanding when such specialization outperforms a strong general-purpose solver, and which design mechanisms drive the gain, is a promising direction. More broadly, applying the LLaMEA framework to other algorithm families---such as multi-objective surrogate-assisted evolutionary algorithms or meta-heuristics---remains a promising direction. Finally, a systematic comparison of LLaMEA against other LLM-driven automated algorithm-design methods, under a common benchmarking protocol, is a natural next step toward understanding the relative strengths of different generation paradigms.

\section{Conclusion}\label{sec:conclusion}

In this work we applied LLaMEA to the automatic generation of multi-objective Bayesian optimization algorithms, showing that large language model-driven evolutionary search can discover competitive and computationally efficient algorithms without any hand-crafted design. Across nine evolutionary runs spanning three evolution strategies, LLaMEA produced approximately $900$ candidates covering fundamentally different surrogate and acquisition-function designs. The development-found \textit{MOEAD-EI Hybrid} attained the highest mean normalized hypervolume on the twelve synthetic benchmark problems (0.971)---significantly more accurate than the state-of-the-art \textit{qParEGO} on 7 of 12 and never significantly worse---while running on average $60\times$ faster; it also generalizes to the real-world problems (0.926, third overall, $\approx$$21.5\times$ faster). On the held-out real-world RE problems, the systematically generated \textit{Improved-Scalarized-EI} was the single most accurate method, significantly outperforming \textit{qParEGO} on two of the three problems at $3.4\times$ lower wall-clock cost. Together these results show that LLaMEA can identify structurally sound algorithm designs that match or exceed state-of-the-art accuracy at a fraction of the computational cost---a balance difficult to achieve through manual design alone.

\bibliography{lib}

\begin{appendices} 

\renewcommand{\thesubsection}{\Alph{subsection}}

\makeatletter
\renewcommand{\p@subsection}{}
\makeatother

\section*{Appendix}

\subsection{Selected LLaMEA-Generated Algorithms}
\label{app:algorithms}

The full source code of all four algorithms is publicly available in the
supplementary repository at:
\begin{center}
\url{https://github.com/glatq/LLaMEA-MOO/tree/publication_llamea_moo_unconstrained#}
\end{center}

\begin{algorithm*}[t]
\caption{Improved-Scalarized-EI}
\label{alg:isei}
\begin{algorithmic}[1]
\Require budget $B$, dimension $d$, bounds $[\mathbf{l},\mathbf{u}]$, window $N_w$,
         weight count $N_\lambda$, EA population $P$, EA generations $G$
\Ensure  non-dominated archive $(\mathcal{F}^\star,\mathcal{X}^\star)$
\State $\mathcal{X}\gets$ Sobol sample of $n_{\mathrm{init}}$ points;\quad
       $\mathcal{F}\gets\textproc{Evaluate}(f,\mathcal{X})$;\quad infer $M$
\State pre-generate $N_\lambda$ uniform weight vectors $\{\boldsymbol{\lambda}_k\}$
       on the simplex (Sobol)
\While{$n_{\mathrm{evals}}<B$}
  \State fit one GP (Mat\'ern-$2.5$) per objective on the last $N_w$ points,
         with objectives normalized to $[0,1]$; on failure, refit without
         optimizer restarts
  \State $\boldsymbol{\lambda}\gets\boldsymbol{\lambda}_{(n_{\mathrm{evals}}\bmod N_\lambda)}$
         \Comment{cycle through weights}
  \State $g^{\min}\gets\min_{\mathbf{y}\in\mathcal{F}}\boldsymbol{\lambda}^\top\tilde{\mathbf{y}}$
         \Comment{best scalarized value (normalized)}
  \State initialize EA population by Sobol sampling, injecting up to $20\%$ of
         the current Pareto points
  \For{generation $=1,\dots,G$}
    \For{each individual $\mathbf{x}$}
      \State predict $(\boldsymbol{\mu},\boldsymbol{\sigma})$ with the GPs;
             scalarize $\mu_s=\boldsymbol{\lambda}^\top\boldsymbol{\mu}$,
             $\sigma_s=\sqrt{\textstyle\sum_m\lambda_m^2\sigma_m^2}$
      \State fitness $\gets\mathrm{EI}(\mu_s,\sigma_s;g^{\min})$
             \Comment{Expected Improvement, minimization}
    \EndFor
    \State elitism (retain best); reproduce by tournament selection, SBX
           crossover and Gaussian mutation; clip to bounds
  \EndFor
  \State $\mathbf{x}^{+}\gets$ best individual found by the EA
  \State $(\mathbf{x}^{+},\mathbf{y}^{+})\gets\textproc{Evaluate}(f,\mathbf{x}^{+})$;
         append to archive; update Pareto front
\EndWhile
\State \Return non-dominated subset of the archive
\end{algorithmic}
\end{algorithm*}

\begin{algorithm*}[t]
\caption{RF-LCB-PBI}
\label{alg:rflcbpbi}
\begin{algorithmic}[1]
\Require budget $B$, dimension $d$, bounds $[\mathbf{l},\mathbf{u}]$, candidate
         pool $N_c$, weights $N_w$, trees $T$, window $W$, PBI penalty $\theta$,
         LCB weight $\kappa$, crowding weight $\kappa_{cd}$
\Ensure  non-dominated archive $(\mathcal{F}^\star,\mathcal{X}^\star)$
\State $\mathcal{X}\gets$ Sobol sample of $n_{\mathrm{init}}$;\quad
       $\mathcal{F}\gets\textproc{Evaluate}(f,\mathcal{X})$;\quad infer $M$
\While{$n_{\mathrm{evals}}<B$}
  \State fit one Random Forest ($T$ trees) per objective on the last $W$ points
  \State $\mathcal{C}\gets$ Sobol sample of $N_c$ candidates
  \State predict mean $\boldsymbol{\mu}$ and ensemble std $\boldsymbol{\sigma}$ for
         $\mathcal{C}$; normalize with the ideal/nadir of $\mathcal{F}$
  \State $\mathbf{f}^{\mathrm{LCB}}\gets\boldsymbol{\mu}_{\mathrm{norm}}-\kappa\,\boldsymbol{\sigma}_{\mathrm{norm}}$
         \Comment{lower confidence bound}
  \State draw $N_w$ weight vectors $\{\mathbf{w}\}$ (Sobol, $\sum_m w_m=1$)
  \For{each weight $\mathbf{w}$}
    \State $d_1\gets(\mathbf{w}^\top\mathbf{f}^{\mathrm{LCB}})/\lVert\mathbf{w}\rVert$;\quad
           $d_2\gets\lVert\mathbf{f}^{\mathrm{LCB}}-d_1\hat{\mathbf{w}}\rVert$;\quad
           $\mathrm{PBI}_{\mathbf{w}}\gets d_1+\theta d_2$
  \EndFor
  \State $s(\mathbf{c})\gets\min_{\mathbf{w}}\mathrm{PBI}_{\mathbf{w}}(\mathbf{c})$
         \Comment{best (most optimistic) subproblem}
  \State $s(\mathbf{c})\gets s(\mathbf{c})-\kappa_{cd}\,\overline{\mathrm{dist}}_k(\mathbf{c},\mathcal{F}^\star)$
         \Comment{crowding-distance bonus ($k{=}3$)}
  \State evaluate the candidate with the lowest $s$; append to archive; update Pareto front
\EndWhile
\State \Return non-dominated subset of the archive
\end{algorithmic}
\end{algorithm*}

\begin{algorithm*}[t]
\caption{RF-ParEGO-Batch}
\label{alg:rfparego}
\begin{algorithmic}[1]
\Require budget $B$, dimension $d$, bounds $[\mathbf{l},\mathbf{u}]$, candidate
         pool $N_c$, batch size $q$, trees $T$, window $W$, exploration weight $\beta$
\Ensure  non-dominated archive $(\mathcal{F}^\star,\mathcal{X}^\star)$
\State $\mathcal{X}\gets$ LHS sample of $n_{\mathrm{init}}$;\quad
       $\mathcal{F}\gets\textproc{Evaluate}(f,\mathcal{X})$;\quad infer $M$
\While{$n_{\mathrm{evals}}<B$}
  \State fit one Random Forest ($T$ trees) per objective on the last $W$ points
  \State $\mathcal{C}\gets$ LHS sample of $N_c$ candidates
  \State predict mean $\boldsymbol{\mu}$ and ensemble std $\boldsymbol{\sigma}$ for
         $\mathcal{C}$; normalize with the observed ideal/nadir
  \State $\mathcal{B}\gets\emptyset$
  \For{$i=1,\dots,q$}
    \State draw weight vector $\mathbf{w}\sim\mathrm{Dirichlet}(\mathbf{1})$
           \Comment{fresh scalarization per batch member}
    \State $t_\mu(\mathbf{c})\gets\max_m w_m\,\mu_{\mathrm{norm},m}$;\quad
           $t_\sigma(\mathbf{c})\gets\max_m w_m\,\sigma_{\mathrm{norm},m}$
           \Comment{Tchebycheff}
    \State $a(\mathbf{c})\gets-\big(t_\mu(\mathbf{c})-\beta\,t_\sigma(\mathbf{c})\big)$
           \Comment{minimize scalarized value, LCB exploration}
    \State add to $\mathcal{B}$ the unselected candidate with highest $a$; mark it selected
  \EndFor
  \State evaluate $\mathcal{B}$; append to archive
\EndWhile
\State \Return non-dominated subset of the archive
\end{algorithmic}
\end{algorithm*}

\begin{algorithm*}[t]
\caption{MOEAD-EI Hybrid}
\label{alg:moead_ei_hybrid}
\begin{algorithmic}[1]
\Require initial-design ratio $r_{\mathrm{init}}$, evaluation budget $B$, dimension $d$, bounds $[\mathbf{l},\mathbf{u}]$,
         population size $N_{\mathrm{pop}}$, surrogate-offspring count
         $N_{\mathrm{off}}$, batch size $q$, window size $N_{w}$,
         augmentation constant $\rho$
\Ensure  non-dominated archive $(\mathcal{F}^\star, \mathcal{X}^\star)$
\Statex
\State $n_{\mathrm{init}} \gets \max\!\big(\min(\lfloor B\cdot r_{\mathrm{init}}\rfloor,\, 2d{+}1),\, 1\big)$
\State $N_{\mathrm{pop}} \gets \min(N_{\mathrm{pop}},\, n_{\mathrm{init}})$
       \Comment{cap population by initial-design size}
\State $\mathcal{X} \gets$ Latin Hypercube sample of $n_{\mathrm{init}}$ points in $[\mathbf{l},\mathbf{u}]$
\State $\mathcal{F} \gets \textproc{Evaluate}(f,\mathcal{X})$;\quad infer $M$ (number of objectives) from first call
\State initialise archive $(\mathcal{X}_{\mathrm{all}},\mathcal{F}_{\mathrm{all}})\gets(\mathcal{X},\mathcal{F})$ and population $(\mathcal{X}_{\mathrm{pop}},\mathcal{F}_{\mathrm{pop}})\gets(\mathcal{X},\mathcal{F})$
\State generate $N_{\mathrm{pop}}$ weight vectors $\{\mathbf{w}_j\}$ (uniform grid if $M{=}2$, Dirichlet samples if $M{>}2$)
\State update ideal point $\mathbf{z}^{*}$ and nadir point $\mathbf{z}^{\mathrm{nad}}$ from $\mathcal{F}_{\mathrm{all}}$
\Statex
\While{$n_{\mathrm{evals}} < B$}
  \State fit one GP surrogate per objective on the last $N_{w}$ points of $(\mathcal{X}_{\mathrm{all}},\mathcal{F}_{\mathrm{all}})$
  \State generate $N_{\mathrm{off}}$ offspring from $\mathcal{X}_{\mathrm{pop}}$ via SBX crossover and polynomial mutation
  \State predict $(\boldsymbol{\mu},\boldsymbol{\sigma})$ for every offspring with the GPs
  \State normalise $\boldsymbol{\mu},\boldsymbol{\sigma}$ and $\mathcal{F}_{\mathrm{all}}$ to $[0,1]$ using $\mathbf{z}^{*},\mathbf{z}^{\mathrm{nad}}$
  \For{each subproblem $j = 1,\dots,N_{\mathrm{pop}}$}
    \State $g^{\min}_j \gets \min$ augmented-Tchebycheff value of $\mathcal{F}_{\mathrm{all}}$ under $\mathbf{w}_j$
    \State compute Expected Improvement $\mathrm{EI}_{j}(\cdot)$ of each offspring on the scalarised subproblem $\mathbf{w}_j$
  \EndFor
  \State for each subproblem pick the EI-maximising offspring; take the union and keep the top $q$ by EI
  \State $(\mathcal{X}_{\mathrm{new}},\mathcal{F}_{\mathrm{new}}) \gets \textproc{Evaluate}(f, \text{selected batch})$
  \State append $(\mathcal{X}_{\mathrm{new}},\mathcal{F}_{\mathrm{new}})$ to the archive; update $\mathbf{z}^{*},\mathbf{z}^{\mathrm{nad}}$
  \For{each new point $(\mathbf{x},\mathbf{f})$ and each subproblem $j$}
    \If{augmented-Tchebycheff$(\mathbf{f},\mathbf{w}_j) <$ that of $\mathcal{F}_{\mathrm{pop}}[j]$}
      \State replace population member $j$ with $(\mathbf{x},\mathbf{f})$
    \EndIf
  \EndFor
  \State $(\mathcal{F}^\star,\mathcal{X}^\star)\gets$ non-dominated subset of $(\mathcal{F}_{\mathrm{all}},\mathcal{X}_{\mathrm{all}})$
\EndWhile
\State \Return $(\mathcal{F}^\star, \mathcal{X}^\star)$
\end{algorithmic}
\end{algorithm*}

\subsection{Complete benchmark tables}\label{app:tables}

\begin{table*}[t]\centering
\caption{Synthetic per-problem mean final HV $\pm$ std (part 1). Best per problem in bold; $\dagger$ marks LLaMEA-generated.}
\label{tab:synth_hv_a}
\begin{tabular}{lcccccc}\toprule
Algorithm & ZDT1 & ZDT2 & ZDT3 & ZDT4 & ZDT6 & DTLZ1 \\ \midrule
\multicolumn{7}{l}{\textit{LLaMEA-generated}} \\
Improved-Scalarized-EI$^\dagger$ & $\mathbf{7.456\pm0.013}$ & $7.702\pm0.004$ & $\mathbf{7.671\pm0.090}$ & $258.380\pm19.173$ & $8.316\pm0.043$ & $1.709\pm3.821$ \\
RF-LCB-PBI$^\dagger$ & $5.027\pm0.216$ & $4.752\pm0.152$ & $5.149\pm0.058$ & $257.637\pm4.908$ & $3.419\pm0.110$ & $244.795\pm179.614$ \\
RF-ParEGO-Batch$^\dagger$ & $4.783\pm0.067$ & $4.665\pm0.205$ & $5.055\pm0.037$ & $255.037\pm7.541$ & $3.285\pm0.171$ & $96.969\pm216.829$ \\
MOEAD-EI Hybrid$^\dagger$ & $7.446\pm0.032$ & $\mathbf{7.753\pm0.038}$ & $7.426\pm0.156$ & $\mathbf{316.766\pm6.394}$ & $\mathbf{8.593\pm0.014}$ & $636.003\pm45.774$ \\
\multicolumn{7}{l}{\textit{SOTA Bayesian optimization}} \\
qParEGO & $6.961\pm0.371$ & $7.210\pm0.368$ & $6.894\pm0.363$ & $273.194\pm10.003$ & $7.019\pm1.272$ & $102.544\pm189.447$ \\
IOC-SAMO-COBRA & $6.605\pm0.010$ & $6.806\pm0.012$ & $6.666\pm0.006$ & $244.274\pm10.257$ & $3.001\pm0.173$ & $74.865\pm163.409$ \\
\multicolumn{7}{l}{\textit{Classical / baseline}} \\
NSGA-II & $5.803\pm0.424$ & $5.864\pm0.616$ & $5.631\pm0.796$ & $291.801\pm14.222$ & $2.894\pm1.153$ & $271.170\pm274.693$ \\
NSGA-III & $5.251\pm1.274$ & $5.655\pm0.133$ & $5.528\pm0.767$ & $294.180\pm15.585$ & $3.896\pm1.715$ & $\mathbf{703.556\pm312.342}$ \\
Random Search & $4.561\pm0.125$ & $4.195\pm0.095$ & $4.843\pm0.117$ & $249.182\pm9.171$ & $2.751\pm0.100$ & $61.605\pm131.617$ \\
\bottomrule\end{tabular}\end{table*}

\begin{table*}[t]\centering
\caption{Synthetic per-problem mean final HV $\pm$ std (part 2).}
\label{tab:synth_hv_b}
\begin{tabular}{lcccccc}\toprule
Algorithm & DTLZ2 & DTLZ4 & DTLZ7 & WFG4 & WFG7 & WFG9 \\ \midrule
\multicolumn{7}{l}{\textit{LLaMEA-generated}} \\
Improved-Scalarized-EI$^\dagger$ & $19.349\pm0.569$ & $0.211\pm0.044$ & $\mathbf{27.187\pm0.146}$ & $2.842\pm0.085$ & $20.983\pm1.116$ & $2.938\pm0.091$ \\
RF-LCB-PBI$^\dagger$ & $22.326\pm0.136$ & $0.384\pm0.153$ & $18.234\pm0.628$ & $2.467\pm0.045$ & $18.110\pm0.464$ & $3.029\pm0.042$ \\
RF-ParEGO-Batch$^\dagger$ & $22.266\pm0.604$ & $\mathbf{0.482\pm0.010}$ & $17.913\pm0.634$ & $2.533\pm0.083$ & $18.906\pm0.410$ & $2.916\pm0.023$ \\
MOEAD-EI Hybrid$^\dagger$ & $25.103\pm0.702$ & $0.409\pm0.289$ & $27.130\pm0.037$ & $\mathbf{3.063\pm0.070}$ & $23.342\pm1.012$ & $3.015\pm0.091$ \\
\multicolumn{7}{l}{\textit{SOTA Bayesian optimization}} \\
qParEGO & $\mathbf{25.631\pm0.216}$ & $0.453\pm0.042$ & $25.579\pm0.971$ & $2.940\pm0.064$ & $\mathbf{23.797\pm1.167}$ & $\mathbf{3.102\pm0.014}$ \\
IOC-SAMO-COBRA & $21.993\pm0.179$ & $0.251\pm0.051$ & $20.866\pm0.131$ & $2.329\pm0.089$ & $17.657\pm0.604$ & $2.725\pm0.067$ \\
\multicolumn{7}{l}{\textit{Classical / baseline}} \\
NSGA-II & $22.449\pm1.976$ & $0.121\pm0.000$ & $21.046\pm2.398$ & $2.813\pm0.205$ & $17.282\pm3.645$ & $2.599\pm0.384$ \\
NSGA-III & $23.495\pm1.007$ & $0.249\pm0.175$ & $20.414\pm1.911$ & $2.579\pm0.294$ & $17.610\pm1.628$ & $2.717\pm0.186$ \\
Random Search & $21.313\pm0.120$ & $0.205\pm0.058$ & $16.218\pm0.429$ & $2.286\pm0.065$ & $17.361\pm0.186$ & $2.771\pm0.041$ \\
\bottomrule\end{tabular}\end{table*}

\begin{table*}[t]\centering
\caption{Mean final hypervolume $\pm$ std across 5 repetitions on the Phase 3 real-world RE problems. Best per problem in \textbf{bold}; $\dagger$ marks LLaMEA-generated.}
\label{tab:re_hv}
\begin{tabular}{lccc}\toprule
Algorithm & RE21 & RE34 & RE37 \\ \midrule
\multicolumn{4}{l}{\textit{LLaMEA-generated}} \\
Improved-Scalarized-EI$^\dagger$ & $\mathbf{35.233\pm0.187}$ & $\mathbf{146.161\pm0.984}$ & $0.534\pm0.009$ \\
RF-LCB-PBI$^\dagger$ & $30.364\pm0.744$ & $124.250\pm0.930$ & $0.474\pm0.008$ \\
RF-ParEGO-Batch$^\dagger$ & $33.177\pm0.220$ & $124.008\pm1.546$ & $0.513\pm0.009$ \\
MOEAD-EI Hybrid$^\dagger$ & $34.596\pm0.119$ & $134.307\pm6.184$ & $0.491\pm0.027$ \\
qParEGO & $34.678\pm0.135$ & $135.598\pm4.970$ & $\mathbf{0.560\pm0.007}$ \\
IOC-SAMO-COBRA & $30.435\pm0.354$ & $111.603\pm3.854$ & $0.424\pm0.015$ \\
\multicolumn{4}{l}{\textit{Classical / baseline}} \\
NSGA-II & $30.727\pm3.460$ & $124.335\pm15.156$ & $0.462\pm0.033$ \\
NSGA-III & $32.557\pm1.438$ & $122.792\pm4.684$ & $0.426\pm0.046$ \\
Random Search & $30.493\pm0.258$ & $108.450\pm4.071$ & $0.437\pm0.015$ \\
\bottomrule\end{tabular}\end{table*}

\subsection{Additional plots}
\label{app:plots}

This appendix contains the complete set of convergence curves and Pareto fronts for all twelve synthetic benchmark problems used in Phases~1 and~2, supplementing the four representative panels shown in the main text (Figs.~\ref{fig:phase1_conv} and~\ref{fig:phase2_conv}). Each curve is the mean over five independent repetitions, with the shaded band denoting $\pm 1\sigma$.

\begin{figure*}[p]
  \centering
  \includegraphics[width=\textwidth]{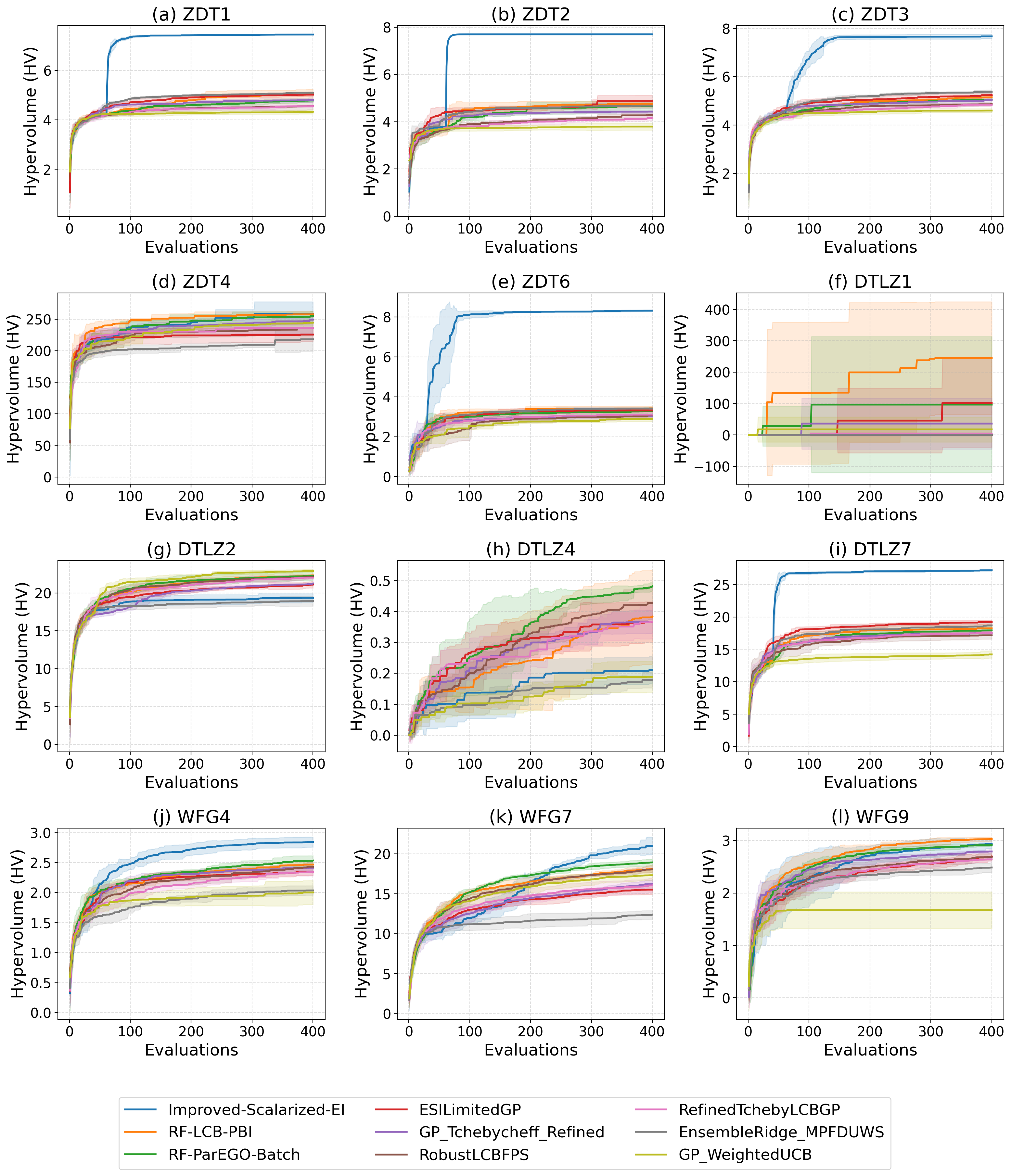}
  \caption{Convergence on all twelve synthetic problems (Phase 1) for the nine
  LLaMEA-generated algorithms. Shaded bands denote $\pm1$ std over five repeats.}
  \label{fig:phase1_all}
\end{figure*}

\begin{figure*}[p]
  \centering
  \includegraphics[width=\textwidth]{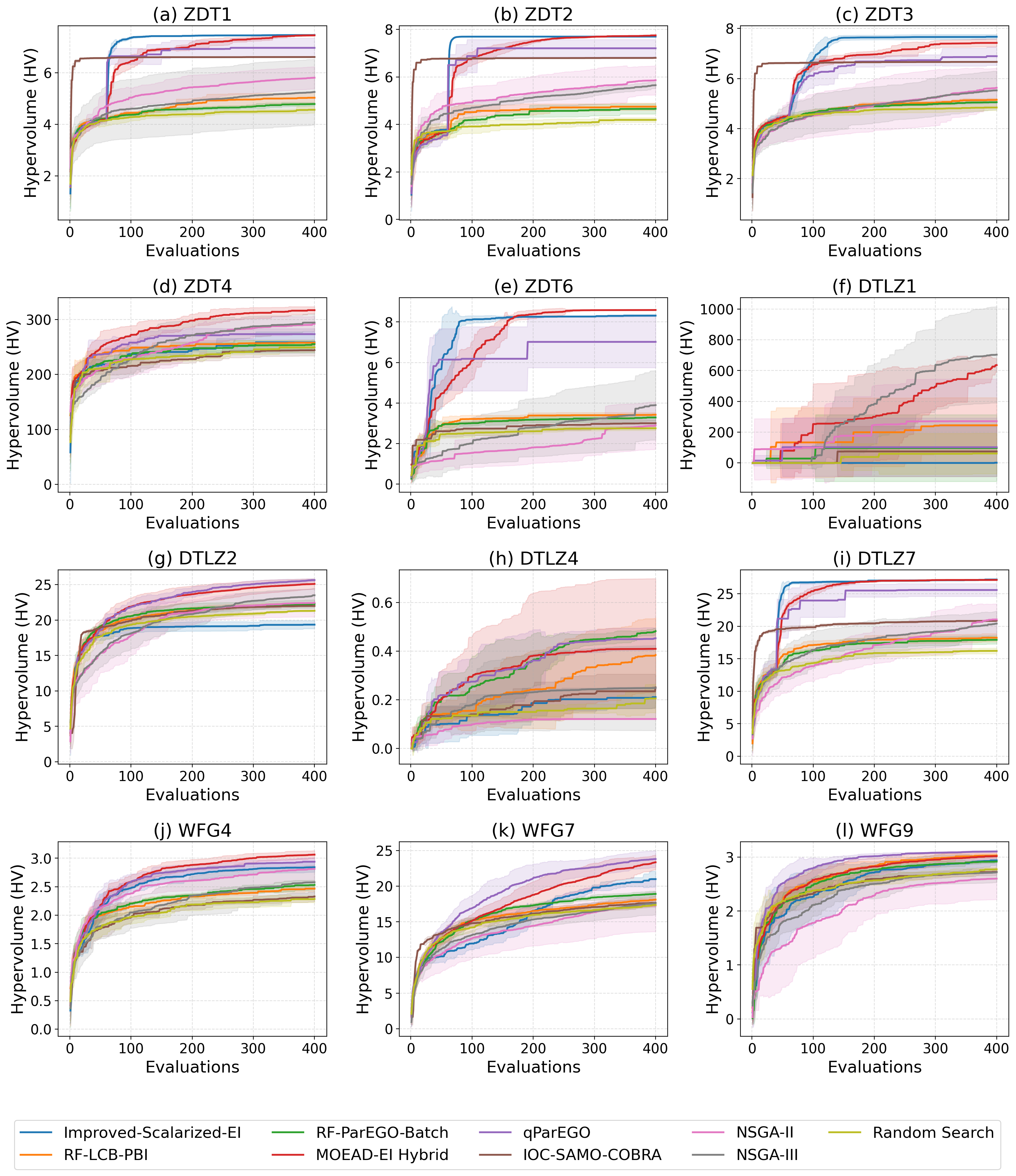}
  \caption{Convergence on all twelve synthetic problems (Phase 2) for the nine benchmarked algorithms (four LLaMEA-generated, the SOTA qParEGO baseline, the surrogate-assisted IOC-SAMO-COBRA, two classical evolutionary baselines, and Random Search). Shaded bands denote $\pm1$ std over five repeats.}
  \label{fig:phase2_all}
\end{figure*}

\begin{figure*}[p]
  \centering
  \includegraphics[width=\textwidth]{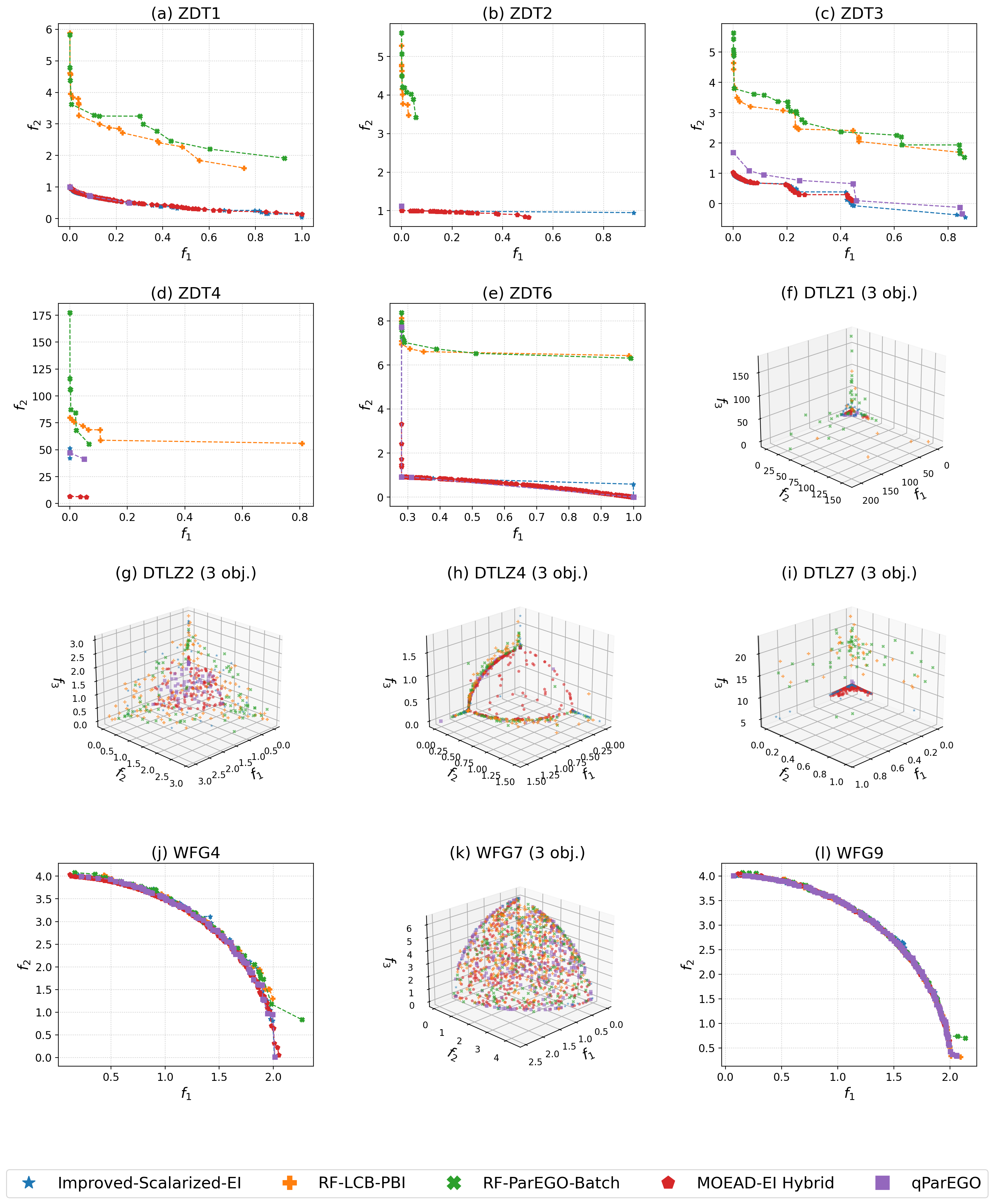}
  \caption{Non-dominated fronts on the twelve synthetic benchmark problems for the
  four LLaMEA-generated algorithms and the SOTA qParEGO baseline. Bi-objective
  problems show the front directly; tri-objective problems (DTLZ1, DTLZ2, DTLZ4,
  DTLZ7, WFG7) are shown as 3D scatters.}
  \label{fig:pareto_synth}
\end{figure*}

\begin{figure*}[t]
  \centering
  \includegraphics[width=\textwidth]{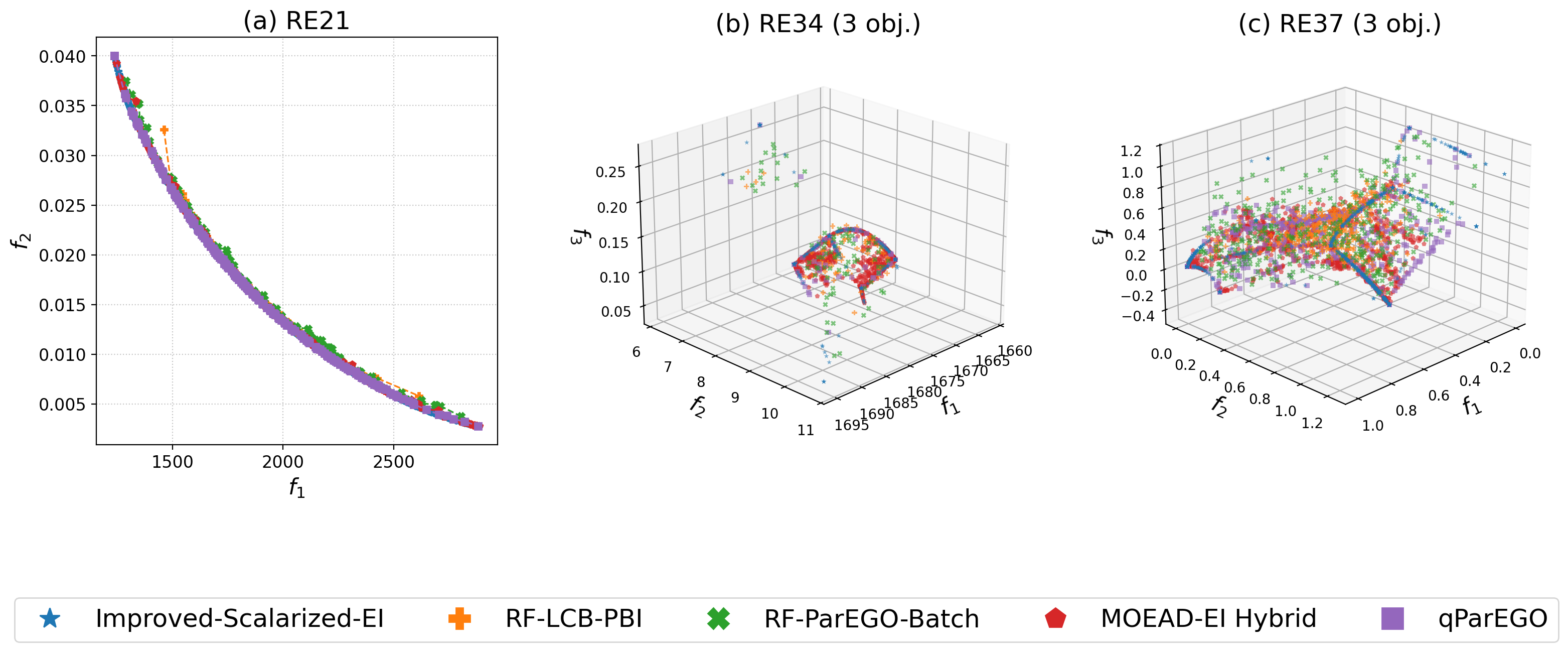}
  \caption{Non-dominated fronts on the real-world RE problems for the four
  generated algorithms and qParEGO. RE21 is bi-objective; RE34 and RE37 are shown
  as 3D scatters.}
  \label{fig:pareto_real}
\end{figure*}

\end{appendices}

\end{document}